\documentclass[lettersize,journal]{IEEEtran}
\usepackage{amsmath,amsfonts}
\usepackage{algorithm}
\usepackage{amssymb}
\usepackage{array}
\usepackage{textcomp}
\usepackage{stfloats}
\usepackage{url}
\usepackage{verbatim}
\usepackage{graphicx}
\usepackage{csquotes}
\usepackage[section]{placeins}
\MakeOuterQuote{"}
\usepackage[%
  bibstyle=ieee,
  citestyle=numeric,
  isbn=true,
  doi=true,
  sorting=none,
  url=true,
  defernumbers=true,
  bibencoding=utf8,
  backend=biber
]{biblatex} 

\DeclareSourcemap{
  \maps[datatype=bibtex]{
     \map{
        \step[fieldsource=doi,final]
        \step[fieldset=isbn,null]
        \step[fieldset=issn,null]
        \step[fieldset=url,null]
        \step[fieldset=urldate,null]
        }
      }
}
\DeclareSourcemap{
  \maps[datatype=bibtex]{
     \map{
        \step[fieldsource=isbn,final]
        \step[fieldset=issn,null]
        \step[fieldset=url,null]
        \step[fieldset=urldate,null]
        }
      }
}

\usepackage{mathtools}
\usepackage[acronym]{glossaries}
\usepackage{tikz}
\usepackage{tikzscale}
\usepackage{mathabx} %
\usepackage[]{xcolor}
\usepackage{etoolbox}
\usepackage{numprint}
\usepackage{xparse}
\usepackage{pgfplotstable} 
\usepackage{tabularx} 
\newcolumntype{C}{>{\centering\arraybackslash}X}
\usepackage[hidelinks,bookmarks=false]{hyperref}
\usepackage[all]{hypcap}    %
\usepackage{orcidlink}

\input{setup/new_commands}
\newacronym{FOT}{FOT}{Field Operational Testing}
\newacronym{SOTIF}{SOTIF}{Safety of the Intended Functionality}
\newacronym{SuT}{SuT}{System under Test}
\newacronym{ODD}{ODD}{Operational Design Domain}
\newacronym{GAMAB}{GAMAB}{Globalement Au Moins Aussi Bon}
\newacronym{GAME}{GAME}{Globalement Au Moins Équivalent}
\newacronym{PRB}{PRB}{Positive Risk Balance}
\newacronym{ALARP}{ALARP}{As Low As Reasonably Possible}
\newacronym{MEM}{MEM}{Minimal Endogenous Mortality}
\newacronym{SBT}{SBT}{Scenario-Based Testing}
\newacronym{PGT}{PGT}{Proving Ground Testing}
\newacronym{AD}{AD}{Autonomous Driving}
\newacronym{ADAS}{ADAS}{Advanced Driver Assistance Systems}
\newacronym{IS}{IS}{Importance Sampling}
\newacronym{AEB}{AEB}{Automatic Emergency Braking}
\newacronym[plural=HBs, firstplural=hazardous behaviors]{HB}{HB}{hazardous behavior}
\newacronym{EVT}{EVT}{Extreme Value Theory}
\newacronym{PoT}{PoT}{Peak over Threshold}
\newacronym{DDT}{DDT}{Dynamic Driving Task}
\newacronym{ASIL}{ASIL}{Automotive Safety Integrity Level}
\newacronym{E/E}{E/E}{Electrical and/or Electronic}
\newacronym{FuSa}{FuSa}{Functional Safety}
\newacronym{ADS}{ADS}{Automated Driving Systems}
\newacronym{V&V}{V\&V}{Verification and Validation}
\newacronym{ENFLI}{ENFLI}{Effectively No Fleet Incidents}
\newacronym{RAR}{RAR}{Risk Acceptance Rationale} 
\newacronym{EVA}{EVA}{Extreme Value Analysis}
\newacronym{TPI}{TPI}{Technical Performance Indicator}
\newacronym{IPM}{IPM}{Intervention Proximity Metric}
\newacronym{PM}{PM}{Proximity Metric}
\newacronym{PDP}{PDP}{Prediction Divergence Principle}
\newacronym{FP}{FP}{False Positive}
\newacronym{FCPr}{FCPr}{False Collision Prediction}
\newacronym{SuS}{SuS}{Subset Simulation}
\newacronym{XiL}{XiL}{X-in-the-Loop}
\newacronym{PD}{PD}{Poisson Distribution}
\newacronym{ED}{ED}{Exponential Distribution}
\newacronym{MTBF}{MTBF}{Mean Time Between Failures}
\newacronym{NATM}{NATM}{New Assessment/Test Method for Automated Driving}
\newacronym{SiL}{SiL}{Software-in-the-Loop}
\newacronym{NHST}{NHST}{Null Hypothesis Significance Test}
\newacronym{SPRT}{SPRT}{Sequential Probability Ratio Test}
\newacronym{QSVRR}{QSVRR}{Quantitative Safety Validation of Residual Risk}
\newacronym{VRU}{VRU}{Vunerable Road User}
\newacronym{TUDa}{TUDa}{Technical University of Darmstadt}
\newacronym{FZD}{FZD}{Institute of Automotive Engineering}
\newacronym{TarA}{TarA}{Testaufwandsreduktion AEB}
\newacronym[plural=VTs]{VT}{VT}{Validation Target}
\newacronym{RP}{RP}{Reduction Path}
\newacronym[longplural={Reduction Approaches}]{RA}{RA}{Reduction Approach}
\newacronym{RC}{RC}{Reduction Cluster}

\pgfplotsset{compat=1.18}
\newcommand{\tikzexternaldisable}{}
\newcommand{\tikzexternalenable}{}
\newcommand{\tikzsetnextfilename}[1]{}
\begin{document}

\title{Towards more efficient quantitative safety validation of residual risk for assisted and automated driving}

\author{%
Daniel Betschinske\,\orcidlink{0000-0003-3203-2296}, 
Malte Schrimpf\,\orcidlink{0000-0002-7661-117X}, 
Steven Peters\,\orcidlink{0000-0003-3131-1664},
Kamil Klonecki, 
Jan Peter Karch\,\orcidlink{0009-0008-7419-836X},
Moritz Lippert\,\orcidlink{0000-0003-2450-0897}

\thanks{Manuscript created in May, 2025; 
The research presented in this paper was conducted at the Institute of Automotive Engineering at the Technical University of Darmstadt in cooperation with and funded by the Continental Autonomous Mobility Germany GmbH.
\textit{(Daniel Betschinske and Malte Schrimpf are co-first authors.)} \textit{(Corresponding authors: Daniel Betschinske; Malte Schrimpf)}%

Daniel Betschinske, Malte Schrimpf, and Steven Peters are with the \acrfull{FZD}, \acrfull{TUDa}, Germany, 64287 Darmstadt (e-mail: firstname.lastname@tu-darmstadt.de).%

Moritz Lippert was with the \acrfull{FZD}, \acrfull{TUDa}. %

Kamil Klonecki und Jan Peter Karch are with Continental Autonomous Mobility Germany GmbH
Guerickestrasse 7, D-60488 Frankfurt, Germany.%

This work has been submitted to the IEEE for possible publication. Copyright may be transferred without notice, after which this version may no longer be accessible.

}
}

\markboth{This work has been submitted to the IEEE for possible publication}%
{\title}
\maketitle
\begin{abstract}
The safety validation of \gls{ADAS} and \gls{ADS} increasingly demands efficient and reliable methods to quantify residual risk while adhering to international standards such as \citenumber{SOTIF2022}. %
Traditionally, \gls{FOT} has been pivotal for macroscopic safety validation of automotive driving functions up to SAE automation level 2. %
However, state-of-the-art derivations for empirical safety demonstrations using \gls{FOT} often result in impractical testing efforts, particularly at higher automation levels. %
Even at lower automation levels, this limitation~\textemdash~coupled with the substantial costs associated with \gls{FOT}~\textemdash~motivates the exploration of approaches to enhance the efficiency of \gls{FOT}-based macroscopic safety validation. %
Therefore, this publication systematically identifies and evaluates state-of-the-art \glspl{RA} for \gls{FOT}, including novel methods reported in the literature.  %
Based on an analysis of \citenumber{SOTIF2022}, two models are derived: a \genericmodel{} capturing the argumentation components of the standard, and a \basemodel{}, exemplarily applied to \gls{AEB} systems, establishing a baseline for the real-world driving requirement for a \gls{QSVRR}.
Subsequently, the \glspl{RA} are assessed using four criteria: quantifiability, threats to validity, missing links, and black box compatibility, highlighting potential benefits, inherent limitations, and identifying key areas for further research. %
Our evaluation reveals that, while several approaches offer potential, none are free from missing links or other substantial shortcomings. %
Moreover, no identified alternative can fully replace \gls{FOT}, reflecting its crucial role in the safety validation of \gls{ADAS} and \gls{ADS}. %
\end{abstract}

\glsresetall
\begin{IEEEkeywords}
     \gls{ADAS}, \gls{ADS}, \gls{FOT}, \citenumber{SOTIF2022}, Residual risk, \acrfull{SOTIF}, Safety validation, Test reduction.
\end{IEEEkeywords}
\glsresetall
\section{Introduction}
\IEEEPARstart{T}{he} automotive industry is constantly striving to increase the safety, comfort, and availability of road vehicles by the implementation of \gls{ADAS} and the increasing automation of the \gls{DDT} as defined by \citenumber{J3016}~\cite{J3016}. %
However, alongside the major challenges involved in developing these new technologies, validating their safety also poses significant difficulties~\cite{Beringhoff.2022, Koopman.2018, Zhao.2023}. %
Ensuring the safety of these systems, defined as the absence of unreasonable risk~\cite[21]{ISO262626P1}, is crucial to gain public trust and regulatory approval. %

To achieve this, both the \textit{microscopic risk} and \macroscopicrisk{} must be considered. %
\textit{Microscopic risk} refers to the risk posed by the system in a single scene~\cite{Junietz.2019} or scenario~\cite{Riedmaier.2020}%
, which is, for instance, determined in \gls{SBT}. %
\textit{Macroscopic risk} refers to the average risk, e.g., the occurrence rate of fatal accidents, posed to society by a new technology or system~\cite{Junietz.2019}. %

Automotive standards such as ISO\,26262~\cite{ISO26262} regarding \gls{FuSa} and \citenumber{SOTIF2022}~\cite{SOTIF2022} addressing the \acrfull{SOTIF}, support system manufacturers with processes, activities, and methods for development, verification, and validation aiming to minimize the \textit{microscopic} and \macroscopicrisk{} of harm emanating from new technologies. %
ISO\,26262 targets the absence of unreasonable risk due to hazards caused by malfunctioning behavior of electrical and electronic systems in road vehicles in general. The scope of \citenumber{SOTIF2022} especially extends the framework for intended functionalities requiring situational awareness obtained by complex sensors and processing algorithms, particularly for emergency intervention systems and driving automation with levels 1 to 5 \cite[1]{SOTIF2022} as defined by SAE\,J3016~\cite{J3016}. %

One of the key activities in the \citenumber{SOTIF2022} framework is the evaluation of \residualrisk{} defined in ISO\,26262 as the "risk remaining after the deployment of safety measures"~\cite[21]{ISO262626P1}. %
This evaluation aims to demonstrate that all potential risks have been identified and mitigated to an acceptable level, providing confidence that the system will operate safely under real-world conditions. %
Thus, the evaluation of \residualrisk{} involves the quantification of the \macroscopicrisk{} posed by the system. %
Traditionally, approaches in the literature address this task by distance-based testing, as various calculation examples show~\cite{Winner.2010b, Kalra.2016, Wachenfeld.2016, Almasri.2024}, but there is no widely accepted agreement about the amount of testing that is generally required~\cite{Singh.2021}. %

Annex\,C3 of \citenumber{SOTIF2022}~\cite{SOTIF2022} includes an exemplary derivation of the real-world driving requirement for an \gls{AEB} system, which serves as a starting point for the considerations of this publication, as outlined in the \basemodel{} in \sectionref{sec::base-model}. %

For \gls{ADS}, i.e., systems with higher automation levels, the current draft of \citenumber{ISO.5083} suggests that: "For L3 automated driving systems a similar amount of real-world test driving as for state-of-the-art L2 systems seems appropriate"~\cite[146]{ISO.5083}. %
However, if the derivation of the real-world driving requirement for higher automation levels is performed in the same way as for L2 systems, the resulting \validationtarget{} for \macroscopicrisk{} evaluation would not be feasible~\cite{Winner.2010b,Kalra.2016}. %

These challenges, coupled with the significant costs associated with \gls{FOT} even at lower automation levels, substantiate the need for more efficient \gls{FOT}.
Motivated by this demand, this work addresses the following research questions:
\begin{enumerate}
    \item Which activities and parameters contribute to the baseline for the real-world driving requirement in \gls{FOT}? %
    \item What approaches are known in the literature or transferable from other fields to enable test effort reduction, and on what grounds is their reduction potential justified? %
\footnote{
As understood in this publication, test reduction does not imply a compromise in safety, but the avoidance of excessive testing by optimized resource use~\textemdash{}~achieving an equal level of confidence with reduced effort.} %
    \item How do the identified approaches for test effort reduction compare in terms of practical applicability, and what are the respective strengths and weaknesses of each approach?
\end{enumerate}

\section{Methodology}\label{sec::Methodology}
\begin{figure}[tbp]
\centering
\tikzsetnextfilename{Fig_1_Methodology}
\includegraphics[]{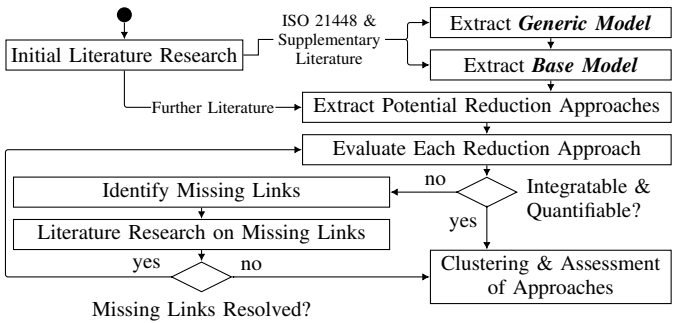}
\caption{Methodology: Identification and Evaluation of Reduction Approaches}
\label{fig::Methodology}
\end{figure}

The methodology used to answer the research questions is shown in \figureref{fig::Methodology}. %
An initial literature research primarily targeting \citenumber{SOTIF2022}~\cite{SOTIF2022} and supplementary sources extracts information relevant for deriving the real-world driving requirement for \gls{FOT}. %
From the results of the literature research, a \genericmodel{} is derived, summarizing key activities and quantities along with their interactions. %
This model serves as a blueprint for the derivation of a real-world driving requirement for \gls{QSVRR}, capturing essential components on a functional level. %
The \genericmodel{} is then instantiated with a minimal practical implementation for the example of an \gls{AEB} system based on Appendix\,C3 of \citenumber{SOTIF2022}, resulting in the \basemodel{}. %
Subsequently, additional literature is reviewed to identify potential \glspl{RA} applicable to the \gls{QSVRR} defined in the \basemodel{}. %
Each approach is evaluated regarding its integrability into the \genericmodel{} and the quantifiability of its effect. %
If integration or quantification is unclear, missing links are documented, and the literature research is iterated accordingly. %
If integration and quantification are possible, the \gls{RA} is visually integrated into the \genericmodel{}, highlighting its precise point of impact and any additional components required for integration. %

\section{Derivation of the Validation Target (VT) for FOT}\label{sec::dev-of-base-model}%
Both the \genericmodel{} and the \basemodel{} are derived from \citenumber{SOTIF2022}. %
The models summarize the fundamental processes and inputs for the derivation of the \gls{VT} that is necessary when performing a \gls{QSVRR}. %
The \gls{VT} is a value to argue that the \textit{acceptance criterion} is met \cite[10]{SOTIF2022}, which represents the criterion for the absence of an unreasonable level of risk \cite[2]{SOTIF2022}. %
While the \genericmodel{} model is not limited to any test method, the \basemodel{} establishes a baseline for the real-world driving requirement  resulting for \gls{FOT}. %
As a starting point for the models, an overview of the structure and the contents of \citenumber{SOTIF2022} is given. %
\subsection[Summary of ISO 21448 Activities]{Summary of \citenumber{SOTIF2022} Activities\protect\footnote{All information in this section is paraphrased or taken directly from \citenumber{SOTIF2022}. Individual citations are therefore not provided for readability.}}\label{sec::SOTIFSummary}%
\begin{figure*}[tbp]
\tikzsetnextfilename{Fig_2_Overview_of_SOTIF_Activities}
    \centering
\includegraphics{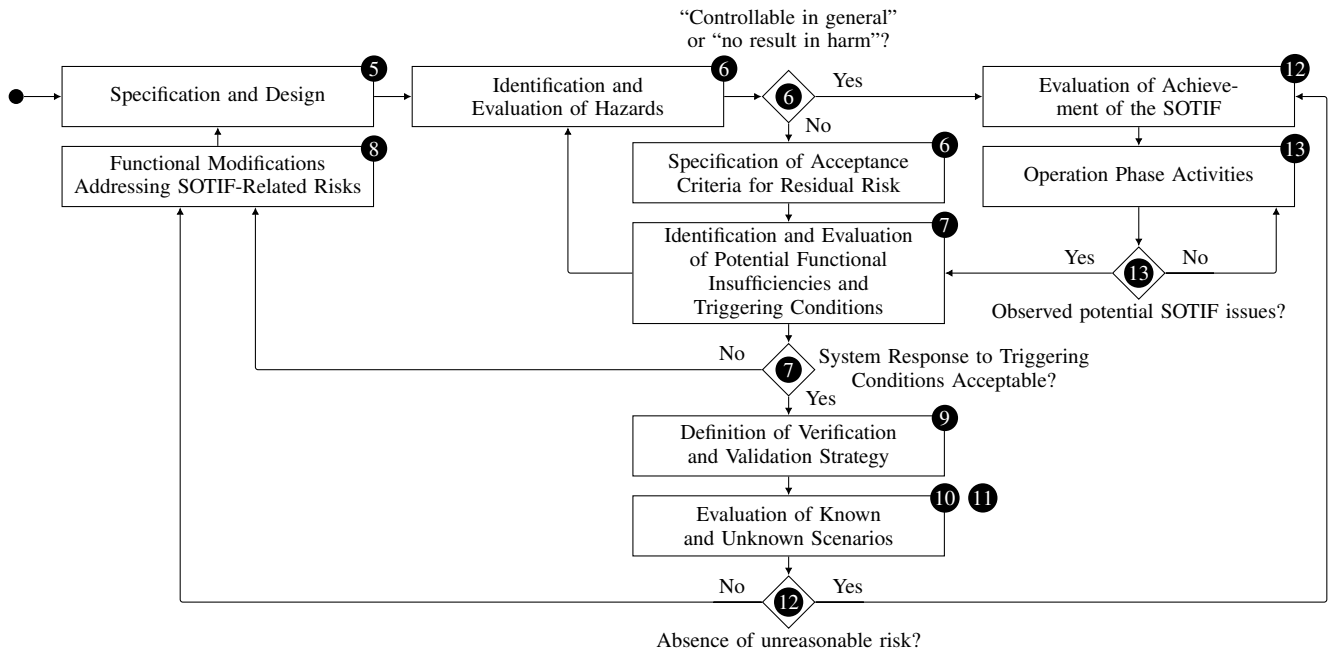}
\caption{Overview of \gls{SOTIF} activities, based on~\cite[18]{SOTIF2022}}
\label{fig::SOTIF_overview}
\end{figure*}

The flowchart depicted in \figureref{fig::SOTIF_overview} provides an overview of the activities associated with the different clauses contained in the \citenumber{SOTIF2022} standard. %
The numbered black circles refer to the number of clauses in \citenumber{SOTIF2022} that describe the corresponding activities. %
In total, \citenumber{SOTIF2022} comprises a set of nine normative clauses that describe activities designed to assure, demonstrate, and maintain the \gls{SOTIF} of systems within its scope across their lifecycle. %
Due to their introductory nature, clauses \SotifClause{1} to \SotifClause{4} are not included in this count. %
\subsubsectionClause{Specification and Design}\label{sec::ClauseFive}%
The activities described in \citenumber{SOTIF2022} begin with the specification and design of a given system, for which the standard provides detailed considerations aimed at supporting the achievement of the \gls{SOTIF}.  %
\subsubsectionClause{a) Identification and Evaluation of Hazards}\label{sec::ClauseSixa}%
Based on a given specification, an analysis is carried out to identify possible hazards arising from the system as specified and to assess associated risks. %
In this context, \textit{hazards} are defined as potential sources of harm caused by \textit{\acrfull{HB}} at the vehicle level, which potentially arise during operation of the \gls{SuT}. %
The \textit{hazards} identified in this manner are evaluated regarding their relevance to \gls{SOTIF} mainly by two criteria: %
It is assessed whether the identified hazards are \textit{controllable in general} or whether the severity of the hazards is generally classified as \textit{no resulting harm}. %
If at least one of these two criteria is met for each \textit{hazard} associated with a specific \gls{HB}, the respective \gls{HB} is not addressed further as it is deemed as not \textit{\gls{SOTIF}-related}~\cite[29]{SOTIF2022}. %
If at least one \textit{hazard} potentially resulting from the identified \glspl{HB} at vehicle level fails both criteria, the respective \gls{HB} is considered \textit{\gls{SOTIF}-related} and is thus further addressed in the following process steps. %
This analysis is explicitly not limited to the ODD of the \gls{SuT}~\cite[43]{SOTIF2022}. %
\subsubsectionClause{b) Specification of Acceptance Criteria for Residual Risk}\setcounter{SOTIFClauseNumber}{6}\label{sec::ClauseSixb}%
If the analysis concludes that a \residualrisk{} remains due to potential \glspl{HB} of the system, \textit{acceptance criteria} for the \residualrisk{} are defined based on the results of the \hyperref[sec::ClauseSixa]{\textit{hazard identification and evaluation}}.%
\subsubsectionClause{Identification and Evaluation of Potential Functional Insufficiencies and Triggering Conditions}\label{sec::ClauseSeven}%
Subsequently, a systematic search for potential insufficiencies in the system specification or implementation is conducted. %
In this context, \textit{insufficiency} refers to a contribution to a \gls{HB} of the system or the failure to detect and/or mitigate potential misuse by the user. %
Furthermore, \textit{triggering conditions} for the aforementioned functional insufficiencies are identified. %
The results  are iteratively linked with the \hyperref[sec::ClauseSixa]{\textit{identification and evaluation of hazards}} and the \hyperref[sec::ClauseSixb]{\textit{specification of associated acceptance criteria}}. %
Based on the analysis results, this process step concludes with a decision on whether the \gls{SOTIF} is deemed achievable without further functional modification.%
\subsubsectionClause{Functional  modifications  addressing \gls{SOTIF}-related  risks}[Functional  modifications  addressing SOTIF-related  risks]\label{sec::ClauseEight}%
If functional modifications are deemed necessary to address \gls{SOTIF}-related risks,  appropriate system modifications shall be implemented by enhancement of system components, functional restrictions, or other appropriate measures. %
These modifications are then incorporated into the system specification and design and the process restarts. %
\subsubsectionClause{Definition of the \gls{V&V} Strategy}[Definition of the V\&V Strategy]\label{sec::ClauseNine}%
If \gls{SOTIF} may be achieved without further functional modifications, a \gls{V&V} strategy with the objective of demonstrating compliance with the previously specified \textit{acceptance criteria} for the \residualrisk{} is developed. %
This includes the specification of the \gls{V&V} methods and corresponding validation objectives. %
The choice of these methods and objectives shall be justified based on their suitability to demonstrate compliance with the \textit{acceptance criteria}. %
\subsubsectionClause{Evaluation of Known Scenarios}
\label{sec::ClauseTen}
The standard defines specific requirements for the evaluation of the previously identified potentially hazardous scenarios as a part of the \gls{V&V} strategy, with a focus on verification activities. %
The recommended methods target the verification of individual system components, such as the sensing subsystem or planning algorithms, and the integrated system as a whole.%
\subsubsectionClause{Evaluation of Unknown Scenarios}\label{sec::ClauseEleven}%
The results of \gls{V&V} activities must also demonstrate that the \residualrisk{} from unknown hazardous scenarios complies with the defined \textit{acceptance criteria}.  %
Recommended test methods include real-world scenario exploration and long-term fleet testing.%
\subsubsectionClause{Evaluation of Achievement of the \gls{SOTIF}}[12 Evaluation of Achievement of the SOTIF]\label{sec::ClauseTwelve}%
Upon completion of the \gls{V&V} activities, the work products generated throughout the \gls{SOTIF} activities are reviewed, and an argument for the achievement of the \gls{SOTIF} shall be provided, serving as the basis for a recommendation for \gls{SOTIF} release. %
\subsubsectionClause{Operation Phase Activities}\label{sec::ClauseThirteen}%
Finally, \citenumber{SOTIF2022} also requires the definition of a field monitoring process to ensure continued compliance with \gls{SOTIF} during the operation phase of the system. %
This includes, for example, monitoring the accuracy of the \residualrisk{} estimate made at the time of release. %
\subsection{The \GenericModel{}}\label{sec::generic_model}%
As the summary in the previous section shows, the activities described in \citenumber{SOTIF2022} and the associated work products contain both quantitative and qualitative elements that contribute to an overall safety argumention. %
In the following, the elements required for the argumentative derivation of the \gls{VT} for the \gls{QSVRR} according to \citenumber{SOTIF2022} are extracted. %
The findings are summarized in the \genericmodel{} shown in \figureref{fig::Generic_Model}. %
The numbered white circles refer to its corresponding components. %
\begin{figure*}[tbp]
\tikzsetnextfilename{Fig_3_The_Generic_Model_of_QSVRR}
\centering
\includegraphics{pictures/Base_Model/Generic_Model.tikz}
\caption{The \GenericModel{} of \gls{QSVRR}}
\label{fig::Generic_Model}
\end{figure*}
\subsubsectionGenericModel{Hazard Identification and Evaluation}\label{sec::HI_Generic}%
Under the assumption that a system specification is given, the entry point for the \genericmodel{} is the identification and evaluation of hazards according to clause 6 of~\citenumber{SOTIF2022} as discussed in \SotifClauseref{sec::ClauseSixa}~\textit{a)}. %
The Output of this process is the identification of one or more \gls{SOTIF}-related \glspl{HB}. %
Within our models, we outline the \gls{QSVRR} for one specific \gls{HB}. If more than one \gls{HB} is identified, all subsequent steps must be performed separately for each \gls{HB}. %
\subsubsectionGenericModel{Benchmark Selection}\label{sec::BES_Generic}%
\citenumber{SOTIF2022} lists several \hypertarget{link::RAR_Generic}{\glspl{RAR}} that may serve as the basis for defining an \textit{acceptance criterion}~\cite[p. 30f]{SOTIF2022}, including: %
\begin{itemize}
    \item The \gls{ALARP} principle, 
    \item The \gls{MEM} principle,
    \item The \gls{PRB} principle, and
    \item \gls{GAME} also known as \gls{GAMAB}.
\end{itemize}
Other \glspl{RAR} are permitted; however, the only alternative that we are aware of is \gls{ENFLI}, proposed by \citeauthor{Stellet.2023c}~\cite{Stellet.2023c}. %
These \glspl{RAR} all are quantitative~\cite{Putze.2023b} and require a benchmark against which risk is compared to justify its acceptance. %
In the \textit{benchmark selection} process, traceable benchmarks are selected and documented for each \gls{HB} in the context of the chosen \gls{RAR}. %
In this context, the benchmark can be understood as the generic event or quantity from which the \textit{acceptance criterion} is derived. %
\subsubsectionGenericModel{Derivation of Acceptance Criterion}\label{sec::DAC_Generic}%
Based on the selected benchmark, an \textit{acceptance criterion} (\AH{}) for the \residualrisk{} of the \textit{harm} associated with the identified \gls{HB} is derived according to clause 6.5 of \citenumber{SOTIF2022}. %
The appropriate data source for deriving \AH{} depends on the selected \gls{RAR}. %
Options include statistical data on the benchmark within the target market (\gls{GAME}, \gls{MEM}), analyses quantifying the risk reduction achieved by the \gls{SuT} (\gls{PRB}), feasibility assessments regarding the implementation of further risk-reduction measures (\gls{ALARP}) or the \textit{acceptance criteria} of comparable, established systems (\gls{GAMAB}). %
Additionally, relevant legislation and other applicable regulations must be considered~\cite[30]{SOTIF2022}.%
\subsubsectionGenericModel{Derivation of Validation Target(s)}\label{sec::DoVT_Generic}%
The goal of the subsequent process is to ensure and demonstrate that the \gls{SuT} meets \AH{}. %
However, since \textit{acceptance criteria} are agnostic to testing and evaluation methods, the definition of the \gls{V&V} strategy serves two main purposes: %
selecting suitable testing methods and translating \AH{} into one or multiple \glspl{VT} with clearly defined pass/fail criteria. %

A key requirement in this process is that the selected methods must ensure \textit{sufficient coverage of the relevant scenario space}~\cite[42]{SOTIF2022}, encompassing known hazardous scenarios as specified in clause 10~\cite[46]{SOTIF2022}), and unknown scenarios as required by clause 11~\cite[50]{SOTIF2022}. %
Accordingly, the test methods and associated \glspl{VT} shall be selected such that they demonstrate compliance with the \textit{acceptance criteria} for the set of scenarios the \gls{SuT} can reasonably be expected to encounter during its lifetime. %
\subsubsectionGenericModel{Evaluation}\label{sec::TE_Generic}%
With the specification of the test methods and \VTi{}, the process for \gls{QSVRR} according to \gls{SOTIF} is completely formalized. %
The execution of the specified test methods generates corresponding observations, e.g., the number of instances in which the system exhibits \gls{HB}. %
The pass or fail outcome is determined by evaluating the observation against the defined \glspl{VT}. %
\subsection{The \BaseModel{}}\label{sec::base-model}%
Based on the \gls{AEB} system example in the normative clauses and appendices of~\citenumber{SOTIF2022}, an exemplary implementation of the \genericmodel{} is derived,  referred to as the \basemodel{}. This \basemodel{} establishes a quantitative baseline for the \gls{VT} required for the \gls{QSVRR} for an \gls{AEB} system and is visualized in \figureref{fig::Base_Model}. %
The numbered light blue circles refer to the corresponding activities of the \basemodel{}.%
\begin{figure*}[tbp]
\tikzsetnextfilename{Fig_4_Base_Model_of_QSVRR}
\centering
\includegraphics[]{pictures/Base_Model/Base_Model.tikz}
\caption{The \BaseModel{} of \gls{QSVRR}}
\label{fig::Base_Model}
\end{figure*}
\subsubsectionBaseModel{Hazard Identification and Evaluation}\label{sec::HI}%
In Clause 6 of \citenumber{SOTIF2022}, \textit{unintended braking} as the result of a \gls{FP} \gls{AEB} intervention is identified as a \gls{HB} at the vehicle level. %
This behavior can lead to different hazards, such as \textit{collisions from behind} or \textit{vehicle occupants falling}, potentially resulting in harm~\cite[28]{SOTIF2022}. %
Since rear-end collisions caused by unintended braking are neither generally controllable nor inherently harmless, the \gls{HB} \textit{unintended braking} is considered \gls{SOTIF}-related and must thus be addressed further. %
\subsubsectionBaseModel{Benchmark Selection}\label{sec::BES}%
The benchmark selection is derived from the example given in Annex C2.2 of \citenumber{SOTIF2022} since it is not explicitly included in the normative clauses of the standard. %
In this example, the likelihood of \textit{unintended braking} caused by the \gls{AEB} functionality must not exceed the likelihood of rear-end collisions caused by humans~\cite[128]{SOTIF2022} for the \textit{residual risk} to be acceptable:
\begin{equation}
    P_{\mathrm{ha,AEB}}\leq P_{\mathrm{ha,hu}}\label{eq::PRB}
\end{equation}
In the given example, no explicit \gls{RAR} is mentioned. However, \equationref{eq::PRB} could be interpreted as an application of the \gls{PRB} principle, since the \gls{AEB} system is designed to prevent human-induced rear-end collisions and, therefore, must not cause more potential rear-end collisions than it could possibly prevent. %
Thus, in this example, the rate of \textit{rear-end collisions} in the field is selected as the \textit{benchmark}. %
This \textit{benchmark} is conservative in the sense that not every unintended braking event will lead to a rear-end collision. However, this selection simultaneously neglects the fact that \gls{AEB} systems typically cannot prevent all potential rear-end collisions in the field. %
While conservative assumptions can be made to achieve traceable benchmarks, more conservative benchmarks generally require more effort to demonstrate that \AH{} is met. %
\subsubsectionBaseModel{Derivation of Acceptance Criterion}\label{sec::DAC}%
As discussed in \GenericModelref{sec::BES_Generic}, various sources may be used to determine a numerical value for the benchmark. %
Annex C2.2  of~\citenumber{SOTIF2022} suggests deriving the average distance traveled by human drivers between rear end collisions~($B$), using traffic statistics of the total yearly distance traveled by all vehicle~($M$) and the annual number of rear end collisions in the field~($A$)~\cite[131]{SOTIF2022}:
\begin{equation}
    B = \frac{M}{A}
\end{equation}
The example suggests that the upper bound for $M$ and the lower bound for $A$ should be used to obtain a worst-case estimate~\cite[132]{SOTIF2022}.
Generally, traffic statistics serve as a reliable reference. %
If available, existing statistics may be adopted without modification. However, in some cases, further calculations or dedicated surveys may be required. %
The derivation of \AH{} can be further refined by a safety factor $>1$ referenced as $\kappa_1$~\cite[132]{SOTIF2022} and $y$~\cite[137]{SOTIF2022}, which serves to strengthen the argumentation by making \AH{} more conservative than the benchmark. %
Furthermore, a factor $<1$ referenced as $\kappa_2$~\cite[132]{SOTIF2022} aims to adjust for the existence of rear-end collisions due to justified braking maneuvers by human drivers in the statistical reference data based on estimations by, e.g., simulations. %
In line with our benchmark selection argumentation based on a \gls{PRB}, an additional factor may be introduced to reflect the fact that \gls{AEB} systems typically do not prevent all potential rear-end collisions in the field.%
\subsubsectionBaseModel{Statistical Model}\label{sec::SMHBA}%
If a statistical test is used to determine if \AH{} is met, it is necessary to assume the underlying distribution of the relevant \gls{HB}. %
For this purpose, the \gls{PD} is selected in alignment with Annex C.3 of \citenumber{SOTIF2022} and~\cite{Littlewood.1997}. %
This implies that the chance of two simultaneous events is negligible or impossible, that the expected value of the random number of events is constant and proportional to the size of the sample, and that the probability of events is independent of each other~\cite{OConnor.2016}. %
\subsubsectionBaseModel{Derivation of the Validation Target (VT)}\label{sec::DoVT}%
Annex C.3 provides an example of how \AH{} can be translated into a singular \gls{VT} termed "real-world driving requirement" $\tau$~\cite[137]{SOTIF2022}, which implies that \gls{FOT} was chosen as the test method associated with $\tau$. %
The core idea is to use a hypothesis test to verify compliance with \AH{}. %
Hence, a stopping criterion needs to be defined that states when to halt data collection and make a decision about the hypothesis. %
There are several ways to derive the stopping rule presented in Annex C.3. %
For the \basemodel{}, we select a \gls{NHST}, because by design it limits the type\,I error, i.e., probability of a \gls{SuT} not meeting the specification passing the test by randomness of the observation. %
To do so, the null hypothesis $H_0$ that the \gls{SuT} performs worse than acceptable, and the alternative hypothesis $H_1$ is that the \gls{SuT} is at least as good as acceptable, are formulated: %
\begin{equation}\label{eq:originalHypothesis}
    H_0: R>A_{\mathrm{H}} \qquad H_1: R\leq A_{\mathrm{H}}
\end{equation}
Here, $A_{\mathrm{H}}$ is the maximum acceptable average rate of harm derived from the \gls{RAR} and $R$ is the actual average occurrence rate of the observed \gls{HB} exhibited by the \gls{SuT}. %
Using the cumulative distribution function of the \gls{PD}, the probability of a system worse than specified exhibiting at maximum $k$ events is bounded by: %
\begin{equation}\label{eg:PMFPoisson}
    P(X\leq k) = \sum_{i=0}^{k} \frac{e^{-\lambda}\lambda^i}{i!} \leq \alpha
\end{equation}
with the expected value of the test $\lambda=A_{\mathrm{H}}\cdot \tau_k$. To guarantee the type\,I error is bounded by the significance\,$\alpha$, the real-world driving requirement $\tau_k$ is selected such that \equationref{eg:PMFPoisson} holds. %
The minimum real-world driving requirement is achieved when only allowing $k=0$ observations, leading to the formulation of the stopping criterion for the distance to be driven in \citenumber{SOTIF2022}~\cite[137]{SOTIF2022}\footnote{In Eq.\,(C.7)
of~\cite{SOTIF2022}, the symbol $\alpha$ is used to describe confidence. However, $\alpha$ describes significance in this publication while $1-\alpha$ describes confidence to maintain conformity with \citenumber{ISO3534}\,\cite{ISO3534}.}:
\begin{equation}\label{eq:minimalvalidationtarget}
    \tau_{k=0} = -\frac{\mathrm{ln}(\alpha)}{A_\mathrm{H}}
\end{equation}
Note that according to the constraints mentioned in \GenericModelref{sec::DoVT_Generic}, any justification for the transfer of \AH{} to the real-world driving requirement $\tau$ requires that the applied test method sufficiently covers the relevant scenario space of known and unknown hazardous scenarios. %
Therefore, it must also be argued that the test profile selection for \gls{FOT} meets this requirement, e.g., by adequate route design. %
\subsubsectionBaseModel{Evaluation}\label{sec::TE}%
In the given example, the evaluation of the \gls{VT} results from the observation made in \gls{FOT}. %
If $k=0$ was selected and no \gls{HB} has occurred after the distance $\tau_{k=0}$, the test is considered passed, and thus, the \gls{QSVRR} has been achieved. %
If, however, a \gls{HB} occurs before the end of the test, the test is considered failed. %
Depending on the potential severity of this \gls{HB}, the system must be modified, and a new test must be initiated. %
Using \gls{NHST}, an extension of the test is not allowed as it would inflate the type~I error rate above its nominal level $\alpha$. %

\section{Overview of Reduction Approaches and Assessment Criteria}\label{sec::ReductionClusters}%
This section provides an overview of the identified \glspl{RA} and introduces the criteria used to assess them. %
A detailed examination of each individual \gls{RA} is presented in the subsequent section. %
\subsection{Reduction Approach Overview}%
\begin{figure*}[btp]
    \tikzsetnextfilename{Fig_5_Systematization_of_Reduction_Approaches}
    \centering
    \includegraphics{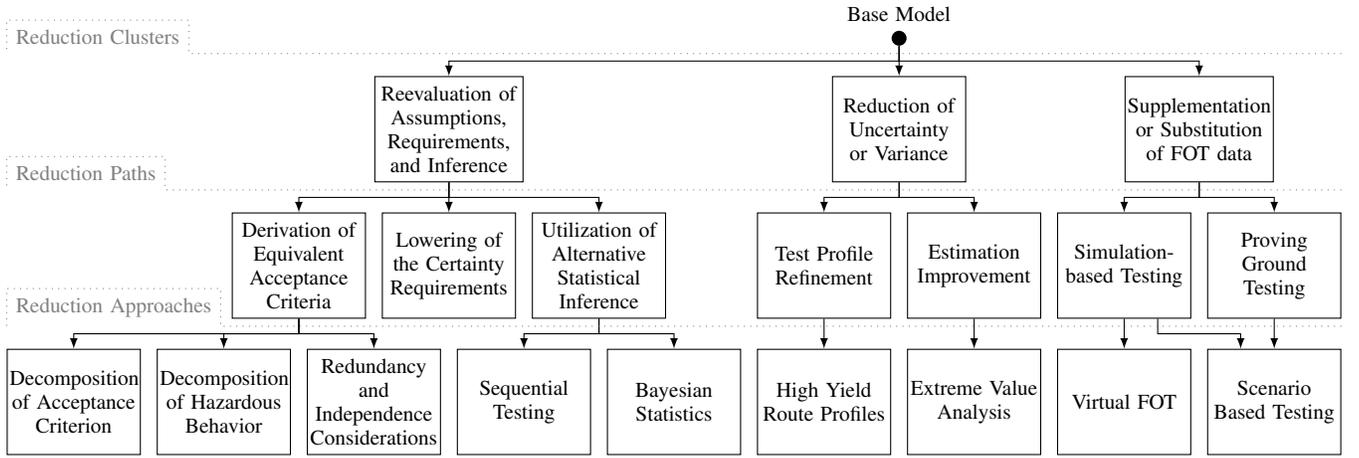}
    \caption{Systematization of Reduction Approaches}
    \label{fig::Reduction_Path_overview}
\end{figure*}

The \glspl{RA} identified are organized into three hierarchical levels to guide the reader as illustrated in \figureref{fig::Reduction_Path_overview}: \glspl{RC}, \glspl{RP}, and \acrfullpl{RA}. %
\glspl{RC} group the approaches by the specific component of the \gls{QSVRR} they target. %
Within \glspl{RC}, \glspl{RP} represent distinct strategies to address an aspect of the \gls{QSVRR}. %
\glspl{RA} are the specific implementations identified in the literature. %
Note that, although \citenumber{SOTIF2022} claims that the "real-world driving requirement can be lessened by using expert knowledge with similar systems"~\cite[137]{SOTIF2022}, expert knowledge is not treated as a standalone \gls{RP} or \gls{RA}. %
Instead, it is considered a supportive element embedded within specific approaches and discussed accordingly. %
More detailed descriptions of the \glspl{RC}, \glspl{RP}, and \glspl{RA} are given in \sectionref{sec::ReductionPaths}. %
\subsection{Assessment Criteria}
Four assessment criteria are defined for evaluating the \gls{RA}: \CritQuantifiablity{}, \CritValidity{}, \CritMissingLinks{}, and \CritBlackbox{}. These criteria are briefly explained in the following subsections and addressed in the discussion of the \glspl{RA}. %
A summary of the evaluations is provided in \tableref{tab:Assessment}, using a three-level rating: positive \Positive, neutral \Neutral, and negative \Negative. %
It is emphasized that these evaluations reflect the subjective perspective of the authors, and no claim of objectivity is made. %
\begin{table*}[tbp]\label{tab:Assessment}
\caption{Summary of Reduction Approach assessments}
\centering
\begin{tabular}{|l||c|c|c|c|}
\hline
\hyperref[fig::Reduction_Path_overview]{\textit{Reduction Approach}}& \CritQuantifiablityShort{} & \CritValidityShort{} & \CritMissingLinksShort{} & \CritBlackboxShort{} \\ \hline
\hyperref[sec::RADecompositionAH]{\RADecompositionAH{}} & \Positive & \Neutral{} ... \Positive{} & \Positive & \Positive
 \\ \hline
\hyperref[sec::RADecompositionHB]{\RADecompositionHB{}} & \Positive & \Neutral~... \Positive & \Neutral{}~... \Positive & \Negative{}~... \Positive
 \\ \hline
\hyperref[sec::RARedundancy]{\RARedundancy{}} & \Positive & \Neutral & \Positive & \Negative
 \\ \hline
\hyperref[sec::RPLoweringRequirements]{\RPLoweringRequirements{}} & \Positive & \Negative & \Positive & \Positive
 \\ \hline
\hyperref[sec::RASequentialTesting]{\RASequentialTesting{}} & \Positive & \Neutral & \Positive & \Positive
 \\ \hline
\hyperref[sec::RABayesianStatistics]{\RABayesianStatistics{}} & \Positive & \Negative{} ... \Positive & \Negative{} ... \Positive & \Positive
 \\ \hline
\hyperref[sec::RAHighYieldRouteProfiles]{\RAHighYieldRouteProfiles{}} & \Negative{}~... \Positive & \Negative{} & \Negative{} & \Neutral{}~... \Positive
 \\ \hline
\hyperref[sec::RAEVA]{\RAEVA{}} & \Positive & \Negative{}~... \Positive & \Neutral & \Negative{} ... \Positive
 \\ \hline
\hyperref[sec::RAVirtualFOT]{\RAVirtualFOT{}} &  \Positive & \Neutral & \Negative & \Negative
 \\ \hline
\hyperref[sec::RASBT]{\RASBT{}} & \Neutral~... \Positive & \Neutral & \Neutral & \Negative~... \Positive
 \\ \hline
\end{tabular}
\end{table*}
\subsubsection{\CritQuantifiablityName{} of the Reduction Approach}\label{sec::CritQuantifiablity}%
This criterion assesses whether the impact of the \glspl{RA} can be measured or quantified. %
An approach is considered more favorable if it provides clear metrics or methods to evaluate its effect on reducing the required \validationtarget{}. %
\begin{itemize}
    \item [\Positive] The approach provides a clear and mathematically described reduction mechanism, allowing for precise quantification of its impact on the \validationtarget{}.
    \item [\Neutral] Parts of the \textit{reduction mechanism} are mathematically describable, but metrics are either unclear or require additional assumptions.
    \item [\Negative] The approach lacks a mathematical description or means to quantify its impact on the \validationtarget{}.
\end{itemize}
\subsubsection{\CritValidityName{} of the Reduction Approach}\label{sec::CritValidity}%
This criterion evaluates aspects introduced by the \gls{RA} potentially compromising the validity of the \gls{QSVRR}. %
Examples include oversimplifying scenarios, omitting critical risk factors, or relying on assumptions that may not hold in real-world applications. %
A lower threat level to the validity indicates a more reliable approach. %
\begin{itemize}
    \item [\Positive] The approach is based on robust and well-documented assumptions that are widely accepted or verifiable. %
    There are no components introducing errors that could compromise the validity of the \gls{QSVRR}. %
    \item [\Neutral] The approach includes moderate threats to validity. %
    Some assumptions or components are not fully validated, but their impact is limited. %
    \item [\Negative] The approach contains significant threats to validity. %
Critical assumptions are unverifiable or weakly justified, and the method includes components where errors, such as flawed expert knowledge, could lead to misleading or invalid results.
\end{itemize}
\subsubsection{\CritMissingLinksName{} of the Reduction Approach}\label{sec::CritMissingLinks}
This criterion evaluates the extent to which a \gls{RA} is ready for application and expresses whether elements still need to be clarified or established to enable practical use. %
It considers whether the necessary data is available or can be collected, whether essential proofs or validations have already been provided, and whether there are unresolved aspects that require further exploration or definition. %
\begin{itemize}
    \item [\Positive] The approach is ready for application. %
    Required data is available, validations are complete, and no significant further work is necessary.
    \item [\Neutral] The approach has potential for application but requires additional steps. %
    Some data needs to be collected, validations of the method or assumptions are still pending, or certain methodological aspects require further definition. %
    \item [\Negative] The approach is not ready for application. %
    Critical data is unavailable or infeasible to collect, essential validations are missing, or major methodological gaps exist that hinder further progress.
\end{itemize}
\subsubsection{\CritBlackboxName{} of the Reduction Approach}\label{sec::CritBlackbox}
This criterion is adopted virtually unchanged from \citeauthor{Riedmaier.2020}~\cite{Riedmaier.2020}, who developed it in the context of an assessment of \gls{SBT}-approaches and formal verification of \gls{ADS}. Our adaptation is limited to a more detailed formulation. The evaluation of the black-box compatibility of an approach expresses the extent to which knowledge of the internal structure of the \gls{SuT} is necessary for its applicability.
\begin{itemize}
    \item [\Positive] A black-box model of the \gls{SuT} is sufficient for an application such that no knowledge of the internal structure of the \gls{SuT} is required. This situation is equivalent to the \basemodel{} (see \sectionref{sec::dev-of-base-model}).
    \item [\Neutral] A gray-box model of the \gls{SuT} is required for an application. This can include, e.g., the code of a software component of the \gls{SuT}, system-internal calculations or the data at an internal interface during runtime. %
    \item [\Negative] The approach cannot be applied without a complete white-box model of the \gls{SuT}. This encompasses at least the complete software of a system, including all interfaces.
\end{itemize}

\section{Reduction Paths}\label{sec::ReductionPaths}
This section discusses the identified \glspl{RC}, \glspl{RP}, and associated \glspl{RA}. %
To guide the reader, the text is organized hierarchically into the three levels introduced in the \hyperref[sec::ReductionClusters]{previous section}. %
\RCSubSection{\RCReevaluation{}}\label{sec::RCReevaluation}
The first cluster focuses on approaches that revisit the foundational premises used to derive the \validationtarget{}. %
This includes redefining \textit{acceptance criteria} and assumptions to achieve a more precise or less conservative \validationtarget{}. %
Thus, it refines the baseline requirements to optimize validation efforts.%
\RPSubSubSection{\RPEquivalentCriteria{}}\label{sec::RPEquivalentCriteria}
Approaches in this \gls{RP} derive equivalent \validationtargets{} by decomposing the top-level criterion into smaller, more manageable subcomponents. %
This decomposition maintains the original safety intent while enabling a more focused and efficient assessment of the system’s compliance with the original \textit{acceptance criterion}.%
\RAParagraph{\RADecompositionAH{}}\label{sec::RADecompositionAH} %
The first \gls{RA} addressed in this publication is the decomposition of the \textit{acceptance criterion} as outlined in Annex C.2 of \citenumber{SOTIF2022}. %
The underlying idea is that in the \basemodel{}, the occurrence rate of \gls{HB} is compared directly with the accepted rate for a certain harm $A_\mathrm{H}$. %
This is a conservative assumption since not every \gls{HB} of the \gls{SuT} will necessarily lead to harm. %
This circumstance can be accounted for using conditional probabilities. %

\citenumber{SOTIF2022} proposes a decomposition that considers exposure, controllability, and severity, analogous to the \gls{ASIL} determination in ISO 26262~\cite{ISO262626P3}. %
This decomposition assumes that there must be exposure to circumstances where the \gls{HB} can cause consequences; those consequences must be uncontrollable by the driver or the system, and the resulting severity must exceed defined thresholds to result in harm with respect to the given \textit{acceptance criteria}. %
The decomposition is formalized using the chain rule, introducing the accepted rate of occurrence at \gls{HB} level $R_{\mathrm{HB}}$: %
\begin{equation}\label{eq:decomposition}
   A_{\mathrm{H}} = P\left(\mathrm{S}|\mathrm{C}\right)\cdot P\left(\mathrm{C}|\mathrm{E}\right)\cdot P\left(\mathrm{E}|\mathrm{HB}\right)\cdot R_{\mathrm{HB}}
\end{equation}
with the respective conditional probabilities:\vspace{3pt}\\
\begin{tabularx}{3.5 in}{>{\raggedright\arraybackslash}p{1.2cm} >{\raggedright\arraybackslash}X}\setlength\tabcolsep{0pt}%
$P\left(\mathrm{E}|\mathrm{HB}\right)$ & Probability of being exposed to circumstances that have the potential for harm in case that the \gls{HB} occurred \\
$P\left(\mathrm{C}|\mathrm{E}\right)$ & Probability that behavior is not controllable in case \gls{HB} occurred and system is exposed to those circumstances\\
$P\left(\mathrm{S}|\mathrm{C}\right)$ & Probability of a certain degree of severity as a result of this uncontrollable situation.
\end{tabularx}\vspace{3pt}
Utilizing \equationref{eq:decomposition}, an acceptable rate for the \gls{HB} can be derived as an equivalent \textit{acceptance criterion} that focuses on the occurrence of the actual \gls{HB} rather than the harm itself:
\begin{equation}\label{eq:RHB}
    R_{\mathrm{HB}}=\frac{A_{\mathrm{H}}}{P\left(\mathrm{S}|\mathrm{C}\right)\cdot P\left(\mathrm{C}|\mathrm{E}\right)\cdot P\left(\mathrm{E}|\mathrm{HB}\right)}
\end{equation}
A detailed discussion of this approach is provided in~\cite{Putze.2023b}. %
\citeauthor{Putze.2023b} highlight that this notation implies purely sequential dependencies, which~\textemdash{}~if inaccurate~\textemdash{}~must be refined by incorporating additional conditional terms within the probabilities. %
They further emphasize that the decomposition inherently assumes a single \gls{HB} leading to the harm. %
However, if multiple \glspl{HB} contribute, their combined effect must be considered by summing their respective probabilities. %
Additionally, \citeauthor{Putze.2023b} note that the values of these conditional probabilities can be system-dependent, particularly for systems with continuous automation, where the \gls{SuT} itself may significantly influence the situations in which it operates. %
To account for these shortcomings,~\citeauthor{Putze.2023b} propose a new notation that addresses these shortcomings and extends the decomposition to the triggering condition.
In more recent literature~\citeauthor{Putze.2023} discusses various methods for determining the respective conditional probabilities~\cite{Putze.2023}. %
Another application of this decomposition is provided in~\cite{Stellet.2023c}. %

The integration into the \basemodel{} for the case of purely sequential dependencies is shown in \figureref{fig::Extension_RADecompositionAH}.
Additional visualization of the more complex approach according to~\citeauthor{Putze.2023b} is analogous, taking into account additional dependencies and conditionals, and thus omitted. %
\begin{figure}[tbp]
\tikzsetnextfilename{Fig_6_Reduction_Approach_Decomposition_of_Acceptance_Criterion}
\centering
\includegraphics[]{pictures/Base_Model/RADecompositionAH.tikz}
\caption{\FigApproachPrefix{}\RADecompositionAH{}}
\label{fig::Extension_RADecompositionAH}
\end{figure}

\textit{Assessment~(see~\tableref{tab:RADecompositionAH}): }
\begin{table}[tbp]
    \centering
    \setlength\tabcolsep{4.5pt}
    \caption{Assessment of \RADecompositionAH{}}
    \begin{tabularx}{\columnwidth}{|c|c|c|C|}
    \hline
    \CritQuantifiablityShort{} & \CritValidityShort{} & \CritMissingLinksShort{} & \CritBlackboxShort{} \\ \hline
    \Positive & \Neutral{} ... \Positive{} & \Positive & \Positive
 \\ \hline
    \end{tabularx}
    \label{tab:RADecompositionAH}
\end{table}
The \gls{RA} \textit{\MakeLowercase{\RADecompositionAH{}}} provides a clear, mathematically defined mechanism to assess its impact on the \validationtarget{} and thus good \CritQuantifiablityshort{}. %
Potential \CritValidity{} arise if dependencies or interactions are neglected during analysis. %
To mitigate these risks, it is essential to verify all assumptions and thoroughly document supporting evidence. %
Furthermore, this decomposition does not incorporate the confidence levels of the individual terms. %
If framed accordingly, this may be acceptable since the resulting values serve as a refined threshold for \glspl{HB}~\textemdash{}~and many established benchmarks do not provide explicit confidence bounds. %
Nevertheless, incorporating confidence levels would strengthen the overall argumentation. %
The approach is otherwise well-documented, requiring no additional steps for application, and is therefore free of \CritMissingLinksshort{}. %
Additionally, the \gls{RA} is independent of system architecture, relying solely on the intended functionality. This makes the approach broadly applicable without requiring in-depth system knowledge and thus \CritBlackbox[black-box compatible]. %

\RAParagraph{\RADecompositionHB{}}\label{sec::RADecompositionHB}%
Unlike the previously described \gls{RA}, which decomposes harm into necessary preconditions to reduce the \validationtarget{}, this approach decomposes \gls{HB} itself. This enables the separate generation of statistical evidence and subsequent aggregation. %
The prerequisite for the approach is a causal chain of effects with one or more events required to trigger the \gls{HB} of a system. %

Originally proposed by \citeauthor{Vaicenavicius.2021}~\cite{Vaicenavicius.2021}, this method has been adopted in Annex C 6.3.2 of \citenumber{SOTIF2022}~\cite[154-156] {SOTIF2022}. %
They define the acceptance rule as follows: Accept if $E[\mathrm{H}]<A_\mathrm{H}$
with confidence ($1-\alpha$), where $E[\mathrm{H}]$ denotes the expected \textit{harm} per distance or time%
\footnote{The notation has been generalized for clarity and differs slightly from~\cite{Vaicenavicius.2021} and Annex C 6.3.2 of~\citenumber{SOTIF2022}.}. %
The equivalent formulation for \gls{HB} is $E[\mathrm{HB}]<R_\mathrm{HB}$, where $R_\mathrm{HB}$ describes the accepted rate for the \gls{HB} derived from $A_{\mathrm{H}}$ (\hyperref[sec::RADecompositionAH]{see RP:~\RADecompositionAH{}}). %
In the example provided by~\citenumber{SOTIF2022}, the \textit{acceptance criterion} $A_{\mathrm{H}}$ is specified in terms of "collisions per distance" due to late or missed detection. 
This specific \gls{HB} of failing to detect an object until a crash with it is unavoidable, can only occur if an object is present. %
Accordingly, the expected rate of \gls{HB} can be decomposed as: %
\begin{equation}\label{eq::HBDecomposition}
     E[\mathrm{HB}] = P(\mathrm{HB}| \mathrm{Trigger}) \cdot E[\mathrm{Trigger}]
\end{equation}
Here, $E[\mathrm{Trigger}]$ is the frequency of the triggering condition (e.g., object presence), and $P(\mathrm{HB}|\mathrm{Trigger})$ is the conditional probability of the system exhibiting the \gls{HB} under that condition. %
Both terms can be estimated sequentially or independent,  each with its own confidence level $(1 - \alpha_1)$ and $(1 - \alpha_2)$. %
If the conditional probability is unknown, it may be approximated using the conditional expectation $E[\mathrm{HB}| \text{Trigger}]$~\cite{Betschinske.2024}. %
The confidence of the aggregated result equals at least $\left(1-(\alpha_1+\alpha_2)\right)$~\cite[p.165]{SOTIF2022}. %

If estimated sequentially, the result of the first estimation (with confidence $(1-\alpha_1)$) may define a derived acceptance threshold $A_{\mathrm{H},2}$ (or $R_{\mathrm{HB},2}$) for the subsequent test. %
Using the upper confidence bound $\hat{E}_{\alpha_1}$ of the first term ensures that the overall \textit{acceptance condition} and total confidence level $(1 - \alpha)$ are preserved. %
\figureref{fig::Extension_HBDecomposition} visualizes this consecutive estimation as extension to the \genericmodel{}. %
\begin{figure*}[!hbtp]
\tikzsetnextfilename{Fig_7_Reduction_Approach_Decomposition_of_Hazardous_Behavior}
\centering
\includegraphics[]{pictures/Base_Model/RADecompositionHB.tikz}
\caption{\FigApproachPrefix{}\RADecompositionHB{}}
\label{fig::Extension_HBDecomposition}
\end{figure*}
In~\cite{Betschinske.2024}, a similar decomposition is applied to \gls{FP} activations of an \gls{AEB} system.
The expected value for \glspl{FP} is expressed as the product of the expected value for \glspl{FCPr} to occur (trigger) and the conditional probability that an \gls{FCPr} causes the decision logic of the \gls{AEB} system to intervene. %
By approximating the probability $P(\mathrm{FP}| \mathrm{FCPr})$ using the upper bound of the conditional expectation $\hat{E}_{\alpha_1}[\mathrm{FP}| \mathrm{FCPr}]$, an equivalent \textit{acceptance criterion} is defined:
\begin{equation}\label{eq:EquivalentTest}
    E[\mathrm{FCPr}]<R_{\mathrm{FCPr}}=\frac{R_{\mathrm{FP}}}{\hat{E}_{\alpha_1}[\mathrm{FP}| \mathrm{FCPr}]}
\end{equation}
The achievement of this \textit{acceptance criterion} can be demonstrated faster than the baseline if \glspl{FCPr} occur significantly more frequently.
An additional benefit of this approach is that it not only derives an equivalent \textit{acceptance criterion} but also provides deeper insights into system performance by analyzing the rate and distribution of \gls{FCPr} generated by the \gls{SuT}. %
However, as shown in simulations in~\cite{Betschinske.2024}, estimating $P(\mathrm{FP}|\mathrm{FCPr})$ from \gls{FOT} data still requires additional samples to meet the desired confidence level. %
A proposed solution is to use input data recordings from legacy systems, provided that the equivalence of the \gls{FCPr} distributions generated on both datasets can be demonstrated. %
If equivalence cannot be established, the estimated conditional expectation may be biased, potentially compromising the validation. %
Conversely, if equivalence of the data sets for the estimation can be demonstrated, or system-independence of the conditional probability can be proven, and if \glspl{FCPr} are indeed more frequent than the original \glspl{FP}, the approach enables significant test reduction potentials, since frequent event estimations converge more quickly. %

\textit{Assessment~(see~\tableref{tab:RADecompositionHB}): }
\begin{table}[tbp]
    \centering
    \setlength\tabcolsep{4.5pt}
    \caption{Assessment of \RADecompositionHB{}}
    \begin{tabularx}{\columnwidth}{|c|c|c|C|}
    \hline
    \CritQuantifiablityShort{} & \CritValidityShort{} & \CritMissingLinksShort{} & \CritBlackboxShort{}\\ \hline
    \Positive & \Neutral~... \Positive & \Neutral{}~... \Positive & \Negative{}~... \Positive
 \\ \hline
    \end{tabularx}
    \label{tab:RADecompositionHB}
\end{table}
The \textit{\MakeLowercase{\RADecompositionHB{}}} offers a mathematically well-defined reduction mechanism resulting in excellent \CritQuantifiablityshort{}. %
A key advantage of this approach is that both estimations may be performed using different methods, enabling individually optimized estimation processes. %
As with \hyperref[sec::RADecompositionAH]{\MakeLowercase{\RADecompositionAH{}}}, neglecting dependencies or interactions during analysis may compromise validity. %
To mitigate this \CritValidity[threat to validity]{}, all assumptions must be carefully validated and thoroughly documented. %
A major limitation arises when a system dependence of the estimated probabilities can not be excluded, as no established method exists to adjust values derived from earlier systems accordingly. %
The approach in~\cite{Betschinske.2024} shifts that issue by estimating the conditional expectation with the current software of the \gls{SuT} while using input data from previous system generations. %
As a result, it must be demonstrated that the use of this additional dataset still yields a valid estimation. %
This, however, states a missing link of the approach that is to be investigated in the future. 
Thus, the \CritMissingLinksshort{} and the \CritBlackbox{} mainly depend on the event and decomposition chosen.%

\RAParagraph{\RARedundancy{}}\label{sec::RARedundancy}
Leveraging redundancy has significant potential to reduce testing effort, as discussed in \citenumber{SOTIF2022} in Annex~C6.3.3~\cite[156-158]{SOTIF2022}: %
Consider a system architecture that includes redundant function channels, each independently implementing a specific subfunction. %
If each channel can perform this subfunction independently, and the \gls{SOTIF} is ensured as long as at least one channel operates correctly, then redundancy can be leveraged to derive reduced \validationtargets{}. %
To do so, the system-level probability of \gls{HB} is decomposed into probabilities of one or more \glspl{HB} for each individual channel. %
This is represented by the component \textit{modeling of redundancy in system architecture} in \figureref{fig::RARedundancy}.
\begin{figure*}[btp]
\tikzsetnextfilename{Fig_8_Reduction_Approach_Redundancy_and_Independence_Considerations}
\centering
\includegraphics[]{pictures/Base_Model/RARedundancy.tikz}
\caption{\FigApproachPrefix{}\RARedundancy{}}
\label{fig::RARedundancy}
\end{figure*}

This decomposition enables the individual \textit{derivation of validation targets} per redundant channel, potentially reducing the validation effort.
However, this approach necessitates a thorough decomposition of potential \glspl{HB}, ensuring that redundancies are applied only when there are no shared failure modes among subcomponents, such as identical functional deficiencies. %
Thus, an \textit{analysis of independence of channels} must be conducted to verify independence between the channels or account for them in the statistical models.
An example of the application of this approach is shown in~\cite{Berk.2017, Berk.2019} in which requirements for the individual sensors are derived from a requirement at the perception level. 
\citeauthor{Berk.2019b} suggest that when redundancy is implemented through majority voting or merging algorithms, k-out-of-n models may be considered as statistical approach for modeling such systems. %
Furthermore, two potential statistical models are employed to model the statistical dependencies: the binomial model, which assumes statistical independence between sensor errors, and the beta-binomial model, which takes sensor error dependence into account.%

\textit{Assessment~(see~\tableref{tab:RARedundancy}): }
\begin{table}[b]
    \centering
    \setlength\tabcolsep{4.5pt}
    \caption{Assessment of \RARedundancy{}}
    \begin{tabularx}{\columnwidth}{|c|c|c|C|}
    \hline
    \CritQuantifiablityShort{} & \CritValidityShort{} & \CritMissingLinksShort{} & \CritBlackboxShort{}\\ \hline
    \Positive & \Neutral & \Positive & \Negative
 \\ \hline
    \end{tabularx}
    \label{tab:RARedundancy}
\end{table}
This approach allows for a \CritQuantifiablity[quantifiable]{} reduction using statistical models. %
However, assessing the overall confidence in meeting the system-level target may be challenging due to the complexity of aggregating individual validation activities for independent channels. %
\CritValidity[Threats to validity]{} mainly arise if dependencies or shared failure modes among channels are insufficiently addressed, making thorough documentation of the independence analysis and statistical modeling critical. %
While the approach does not have any major \CritMissingLinksshort{}, its effectiveness is highly system-dependent, as the level of independence and redundancy is influenced by specific architecture and sensor design choices. %
This limits its general applicability. %
For practical application of the approach, a white-box model  of the \gls{SuT} should be provided since simplification may increase the chance of neglecting dependencies, which justifies the negative rating for the \CritBlackbox{}. %
Furthermore, in most cases, the approach is only applicable to subfunctions as implementing a fully independent channel for a complex function like \gls{AEB} is typically neither economically feasible nor desirable. %

\RPSubSubSection{\RPLoweringRequirements{}}\label{sec::RPLoweringRequirements}
The most trivial \gls{RP} is to lower the required test distance by relaxing the necessary confidence or the refinement factors discussed in \BaseModelref{sec::DAC} and explicitly visualized in \figureref{fig::Extension_RPLoweringRequirements}. %
While~\citenumber{SOTIF2022} mandates a minimum safety factor of 1, the standard does not define a requirement for the confidence level. %
\begin{figure}[b]
\tikzsetnextfilename{Fig_9_Reduction_Approach_Lowering_of_the_Certainty_Requirements}
\centering
\includegraphics[]{pictures/Base_Model/RPLoweringRequirements.tikz}
\caption{\FigApproachPrefix{}\RPLoweringRequirements{}}
\label{fig::Extension_RPLoweringRequirements}
\end{figure}
Due to the logarithmic impact of $\alpha$ in \equationref{eq:minimalvalidationtarget}, the real-world driving requirement increases rapidly with increasing confidence level. %
For example, achieving 99\% confidence instead of the 67\% used in~\citenumber{SOTIF2022} increases the required distance by a factor of 4.6 which means that not only the minimum test distance but also the demands on the system are greatly increased by high confidence requirements~\cite[77-78]{Wachenfeld.2017}. %
Although this \gls{RP} is very obvious, we could not identify any literature that suggests an implementation of a test reduction based on the \textit{lowering of the certainty requirements}. %
Despite not formally qualifying as a \gls{RA}, it is still assessed and included in \tableref{tab:Assessment}. %

\textit{Assessment~(see~\tableref{tab:RPLoweringRequirements}): }
\begin{table}[bp]
    \centering
    \setlength\tabcolsep{4.5pt}
    \caption{Assessment of \RPLoweringRequirements{}}
    \begin{tabularx}{\columnwidth}{|c|c|c|C|}
    \hline
    \CritQuantifiablityShort{} & \CritValidityShort{} & \CritMissingLinksShort{} & \CritBlackboxShort{}\\ \hline
    \Positive & \Negative & \Positive & \Positive
 \\ \hline
    \end{tabularx}
    \label{tab:RPLoweringRequirements}
\end{table}
Since this \gls{RP} is fairly simple and universally applicable, it gets a positive rating concerning \CritQuantifiablityshort{}, \CritMissingLinksshort{}, and \CritBlackbox{}. %
With a sound argument, justifying the change of the safety margin or the confidence might be permissible. %
However, since confidence levels and the refinement factors ultimately quantify the uncertainty of the test, a reduction without good justification poses a major \CritValidity[threat to validity]{}. %
An unjustified reduction may be interpreted as negligence and contradicts the general safety principle \gls{ALARP}.%
\footnote{The socio-ethical problem of initially selecting those parameters is not further discussed within this publication.}%

\RPSubSubSection{\RPAlternativeInference{}}\label{sec::RPAlternativeInference}%
Statistical inference comprises two main branches: estimation and hypothesis testing~\cite{Casella.2002}. %
Both are applicable for evaluating whether the \gls{SuT} likely meets a specified \textit{acceptance criterion}. %
Estimation~\textemdash{}~defined in \citenumber{ISO3534} as "the procedure that obtains a statistical representation of a population from a random sample"~\cite{ISO3534}~\textemdash{}~is primarily descriptive. %
In contrast, a statistical test~\textemdash{}~defined as "the procedure to decide if a null hypothesis is to be rejected in favor of an alternative hypothesis"~\cite{ISO3534} ~\textemdash{}~typically aims to control decision errors (frequentist) or makes a decision based on how much more likely one hypothesis is compared to another (Bayesian)~\cite{Linden.2014}.
Type~I and type~II errors denote incorrect rejections of the null and the alternative hypothesis, respectively. %

As noted in \BaseModelref{sec::DoVT} the \basemodel{} uses \gls{NHST}, a widely used method in frequentist statistics~\cite{EmmertStreib.2024} with predefined sample size and \textit{acceptance criteria}.  %
In \gls{NHST}, the p-value~\textemdash{}~i.e., the probability of observing the test result or a more extreme one under the null hypothesis~\textemdash{}~is constrained by the significance level $\alpha$, thereby bounding the type~I error. %
Note that confidence intervals yield equivalent conclusions, but only when computed with matching predefined sample sizes and significance levels~\cite{Hespanhol.2019}. %

Despite its widespread use, \gls{NHST} is subject to criticism due to frequent misinterpretations and unaccounted limitations~\cite{Greenland.2016}.
Neglecting these limitations can lead to significant deviations from postulated type\,I and type\,II error rates~\cite{Yu.2014}. %
Those practices, whether executed intentionally or unintentionally, are often referred to as different types of \textit{p-hacking}~\cite{Stefan.2023}.
An example is the \textit{optional stopping} of a \gls{NHST}, which describes the practice of extending the test duration if the initially defined significance or maximum number of events could not be achieved by the end of the test. %
Even if early results strongly suggest that the system meets the requirement, continuing the test grants a \gls{SuT}~\textemdash{}~which might actually perform worse on average than the requirement~\textemdash{}~another opportunity to pass due solely to chance, ultimately inflating the type\,I error~\cite{Stefan.2023,Armitage.1969}. %
To address these issues, the following sections introduce alternative statistical inference methods from the literature and examine their potential to reduce the real-world driving requirement.

\RAParagraph{\RASequentialTesting{}}\label{sec::RASequentialTesting}
Sequential testing is a frequentist statistical method that enables continuous evaluation of hypotheses as data is collected. %
A prominent example is Wald's \gls{SPRT}~\cite{Wald.1945}, which has been proposed for \gls{QSVRR} for \gls{ADAS} in~\cite{Almasri.2024}. %
To check whether a test should be continued or stopped in favor of the null or an alternative hypothesis, the \gls{SPRT} calculates the likelihood ratio of the observed data under the null and alternative hypothesis. %

However, this test method is not without its own limitations. %
To apply the \gls{SPRT} approach as described in~\cite{Almasri.2024}, the null and alternative hypotheses must be defined as simple hypotheses. E.g.: %
$$H_0: \theta=\theta_0 \qquad H_1: \theta=\theta_1$$
where $\theta$ is the \gls{MTBF} of the \gls{SuT} and $\theta_0$ and $\theta_1$ are the \gls{MTBF} of the null and alternative hypotheses. %
In~\cite{Almasri.2024}, $\theta_0$ represents the acceptable \gls{MTBF} derived from observational data. The alternative hypothesis, $\theta_1$, is determined using a Chi-Squared confidence interval in conjunction with $\theta_0$, resulting in $\theta_1 < \theta_0$. %
To our knowledge, this derivation of the alternative hypothesis is arbitrary but has significant implications for the derived test efforts, as discussed in the following. %
With the assumption of exponentially distributed (ED) distances between events and the total driven distance $d_k = \sum_{i=1}^k d_i$ until the $k$-th observed event, the test continuation region is given by: %
\begin{align}
d_{H_0}(k) < d_k < &d_{H_1}(k) \quad &\text{for} \quad \theta_0 < \theta_1 \label{eq::RASequentialTestingStoppinga}\\
d_{H_0}(k) > d_k > &d_{H_1}(k) \quad &\text{for} \quad \theta_0 > \theta_1 \label{eq::RASequentialTestingStoppingb}
\end{align}
with $k\in\mathbb N_{>0}$, where
\begin{align}
d_{H_0}(k) &= \frac{\log \left( \frac{\theta_0}{\theta_1} \right) \cdot k + \log \frac{1-\alpha}{\beta}}{\frac{1}{\theta_1} - \frac{1}{\theta_0}} \\
d_{H_1}(k) &= \frac{\log \left( \frac{\theta_0}{\theta_1} \right) \cdot k - \log \frac{1-\beta}{\alpha}}{\frac{1}{\theta_1} - \frac{1}{\theta_0}}
\end{align}
If the right side of the inequality in \equationref{eq::RASequentialTestingStoppinga} or \equationref{eq::RASequentialTestingStoppingb} does not hold, the test is stopped in favor of the null hypothesis. %
If the left side of the inequality does not hold, the test is stopped in favor of the alternative hypothesis. %
The quantities $\alpha$ and $\beta$ quantify the type\,I and type\,II error probabilities. %
The acceptance boundary for $H_0$ is therefore a function of the observed failures $k$, with an offset of $(\log \frac{1-\beta}{\alpha})/(\frac{1}{\theta_1}-\frac{1}{\theta_0})$ and a slope of $ \log \left ( \frac{\theta_0}{\theta_1} \right )/(\frac{1}{\theta_1}-\frac{1}{\theta_0})$. %
Thus, the acceptance boundary is not only dependent on the observed failures of the \gls{SuT} and the \textit{acceptance criterion}, but also on the alternative hypothesis $H_1$. %
If the selected $\theta_1$ gets close to $\theta_0$, the slope converges towards $\theta_0$, whereas the offset tends towards infinity, resulting in infeasible validation efforts. %
The integration into the \genericmodel{} is shown in \figureref{fig::RASequentialTesting}. %
\begin{figure*}[btp]
\tikzsetnextfilename{Fig_10_Reduction_Approach_Substitution_of_NHST_by_RASequentialTesting}
\centering
\includegraphics[]{pictures/Base_Model/RASequentialTesting.tikz}
\caption{\FigApproachPrefix{}Substitution of \gls{NHST} by \RASequentialTesting{}}
\label{fig::RASequentialTesting}
\end{figure*}
Note that exact error guarantees only hold when $\theta$ equals either $\theta_0$ or $\theta_1$~\cite{Wald.1945}. %
This is a significant limitation of Wald's \gls{SPRT}, since the \gls{MTBF} of the \gls{SuT} is usually unknown. %
Nevertheless, the test can be used to test the composite hypothesis $H_0: \theta\leq\theta_0$ against $H_1: \theta\geq\theta_1$~\cite{Wald.1945}. %
In such cases, the stopping rule derived for simple hypotheses (see \equationref{eq::RASequentialTestingStoppinga}) still bounds type\,I and II errors by $\alpha$ and $\beta$ as long as both hypothesis are chosen more strictly than the specification~\cite{Wald.1973,Lai.1988}. %
Thus, to ensure that a \gls{SuT} with sub-specification performance passes with at most probability $\alpha$, both $\theta_0$ and $\theta_1$ must be set more conservatively than the acceptable threshold:
\begin{equation}\label{eq:SPRT_Theta}
    \theta_0,\theta_1 \geq\theta_{A_{\mathrm{H}}}
\end{equation}
In~\cite{Almasri.2024}, this requirement is not met, as $\theta_1 < \theta_0 = \theta_{A_{\mathrm{H}}}$. %
As a result, the test decides based on whether the system is more likely to be barely acceptable or worse, resulting in lower real-world driving requirements compared to the \basemodel{} test. %
However, the test does not ensure that all systems performing worse are rejected with probability $\geq 1-\beta$. %

For $\theta_0 = \theta_{A_{\mathrm{H}}}$ and $\theta_1 > \theta_0$, the minimum real-world driving requirement to accept $H_1$ using \gls{SPRT} is $d_{H_1}(k=1)$. To enable a lower real-world driving requirement than the \basemodel{} (see \equationref{eq:minimalvalidationtarget}) with the same type\,I error guarantee, the conditition:
\begin{equation}
    \beta>1-c \quad \text{with}\quad c =\frac{\theta_0}{\theta_1}\alpha^{\theta_0/\theta_1}
\end{equation}
must hold. For $\theta_0,\theta_1\in\mathbb{R}_{>0}$ the value of $c$ is bounded by $c\leq(-ln(\alpha)\cdot e)^{-1}$. %
As a result, $\beta$ has to exceed a value of approximately 92\% for $\alpha=0.01$. %
The downside of selecting $\beta$ to enable such a reduction using the \gls{SPRT} is that even systems meeting the criteria $\theta_1>\theta_0$ would fail with a relatively high probability. %

\textit{Assessment~(see~\tableref{tab:RASequentialTesting}): }
\begin{table}[tb]
    \centering
    \setlength\tabcolsep{4.5pt}
    \caption{Assessment of \RASequentialTesting{}}
    \begin{tabularx}{\columnwidth}{|c|c|c|C|}
    \hline
    \CritQuantifiablityShort{} & \CritValidityShort{} & \CritMissingLinksShort{} & \CritBlackboxShort{}\\ \hline
    \Positive & \Neutral & \Positive & \Positive
 \\ \hline
    \end{tabularx}
    \label{tab:RASequentialTesting}
\end{table}
The biggest advantage of this test method is not the potential to reduce the minimum possible real-world driving requirement, but to allow for optional stopping. %
This aspect offers reduction potential if multiple events occur compared to \gls{NHST}, which would require the test to be restarted on each event exceeding the pass criteria to maintain the type I error guarantee. %
The stopping criterion derived from Wald's \gls{SPRT} is \CritQuantifiablity[quantifiable]{}, and the method is well documented, without any \CritMissingLinksshort{}. %
Wald's \gls{SPRT} is also broadly applicable as a substitute for conventional \gls{NHST} and remains \CritBlackbox[black-box compatible]{}. %
\CritValidity[Threats to validity]{} primarily arise from incorrect application, particularly since the error guarantees hold only if condition \equationref{eq:SPRT_Theta} is satisfied. %
Otherwise, while the test remains valid, it no longer ensures that a system worse than acceptable passes with a maximum probability of $\alpha$, as guaranteed when using \gls{NHST}.%

\RAParagraph{\RABayesianStatistics{}}\label{sec::RABayesianStatistics}%
The \gls{NHST} and Wald’s \gls{SPRT} are rooted in frequentist statistics, which evaluate the likelihood of observed data under a fixed hypothesis~\cite{Hespanhol.2019}. %
This framework assumes that either the null or alternative hypothesis is in fact true, and variation is due to randomness~\cite{Wagenmakers.2008a}. %
To determine whether the null hypothesis should be rejected, the probability of obtaining the observed data (or more extreme results) under the null hypothesis is calculated, which is called a p-value. %
If this p-value is smaller than a predefined significance level, the null hypothesis is rejected, providing evidence favoring the alternative hypothesis~\cite{Linden.2014c}. %
Thus, by design, the test only limits the maximum probability that a system with the chosen specification or worse will pass the test in multiple attempts by chance, but does not answer the question of how good the system actually is. %
The reverse conclusion, that a system that passes the test is better, is generally invalid~\cite{Berk.2017, Greenland.2016, Vidgen.2016}. %

An alternative approach to frequentist statistics is Bayesian statistics~\cite{OConnor.2016,FornaconWood.2022}.
Using Bayesian statistics, a real-world driving requirement can be derived similarly. %
In fact~\cite{Littlewood.1997}, the source of the real-world driving requirement formula in \citenumber{SOTIF2022}~\cite[p.137]{SOTIF2022}, is based on Bayesian statistics. %
Another example of its usage is the derivation of the \validationtarget{} for redundant sensors in~\cite{Berk.2017,Berk.2019,Berk.2019b}. 
In Bayesian statistics, the obtained data (observations) is treated as fixed and used to update the belief about a statistical process. %
The initial belief is expressed by a prior distribution representing the beliefs about a parameter before observing the data, though it can be subjective~\cite{FornaconWood.2022}. %

The entire approach is integrated into the \genericmodel{} in \figureref{fig::RABayesianStatistics}, starting with the \textit{definition of a prior distribution}, which usually consists of two steps. %
\begin{figure*}[bp]\tikzsetnextfilename{Fig_11_Reduction_Approach_Substitution_of_NHST_by_Bayesian_Statistics}
    \centering
    \includegraphics[]{pictures/Base_Model/RABayesianStatistics.tikz}
    \caption{\FigApproachPrefix{}Substitution of NHST by \RABayesianStatistics{}}
    \label{fig::RABayesianStatistics}
\end{figure*}
The selection of a prior distribution type~\textemdash{}~potentially incorporating assumptions about the statistical behavior of the described event~\textemdash{}~and the parametrization. %
Both have to be based on a thoroughly documented \textit{rationale} since the choice of the prior can significantly influence the posterior~\cite{Berger.1990}. %
After observing data, the \textit{posterior distribution}, representing the new knowledge about the parameter, is computed in a \textit{Bayesian update}. %
The expected test distance can be determined by a \textit{convergence analysis} which involves first defining the \textit{expected number of \glspl{HB}} and then calculating the distance at which the test pass criteria are expected to be met. %
The minimum real-world driving requirement results for a number of zero occurrences of \gls{HB}, as is the case in the \basemodel{}. %

The pass criteria in \textit{Bayesian inference} are usually defined by either the \textit{Bayes factor} in \textit{Bayesian hypothesis testing} or by the probability $\gamma$ that the parameter is larger or smaller than specified (depending on the use-case) derived from the posterior distribution after \textit{Bayesian estimation}. %
The \textit{Bayes factor} assesses the likelihood ratio of a hypothesis over an alternative hypothesis given the observed data~\cite{FornaconWood.2022}. %
The probability $\gamma$ is assessed on the estimated \textit{credible interval} of the parameter based on its posterior distribution. %
The examples in~\cite{Littlewood.1997,Berk.2017,Berk.2019,Berk.2019b} all use the latter. %
Furthermore, \citeauthor{Littlewood.1997} as well as \citeauthor{Berk.2017} all assume the \gls{HB} to be Poisson distributed in their derivations and thus select the gamma distribution as conjugate prior. %
For the parametrization, \citeauthor{Berk.2017} use Jeffreys' prior in their analysis for the derivation of the real-world driving requirement, yielding $\mathrm{gamma}(\alpha\mathbin{=}0.5,\beta\mathbin{=}0)$~\cite{Berk.2017,Berk.2019,Berk.2019b}. %
Jeffreys' priors is a widely accepted prior,  which is invariant under reparametrization and typically considered non-informative~\cite{Jeffreys.1946,Tian.2024}. %
It maximizes the expected posterior information from the data, reducing the effect of the prior~\cite{OConnor.2016}. %

Compared to the real-world driving requirement derived with \gls{NHST} in the \basemodel{} or the example from \citenumber{SOTIF2022}, the minimal real-world driving requirement derived using Jeffreys' gamma-prior is smaller, implying a significant potential for test reduction. %
However, compared to a flat prior, Jeffreys' gamma-prior allocates more relative density near zero than a flat prior, contradicting the \textit{principle of indifference}. %

The \textit{principle of indifference} states that if there is no prior knowledge about the parameter, the prior should be chosen to be evenly distributed over the parameter space~\cite{Keynes.2012b}. %
This would be the case for the improper gamma prior with $\mathrm{gamma}(\alpha\mathbin{=}1,\beta\mathbin{=}0)$~\cite{OConnor.2016}, which exactly matches the prior distribution used in derivation of the real-world driving requirement in the \basemodel{} and in~\cite{Littlewood.1997}. %

In the literature, it is highly debated whether \textit{optional stopping} is less of an issue in Bayesian statistics~\cite{Rouder.2014,Deng.2016,Heide.2021}. %
We emphasize that one might agree because the interpretation of Bayesian quantities like the Bayes factor do not depend on the stopping rule per definition~\cite{Rouder.2014}. %
Unlike the \textit{confidence intervals} in frequentist statistics, the \textit{credible intervals} used in Bayesian statistics do indicate the most probable value for the parameter given the data and the prior and not the likelihood of the data given the parameter. %
However, if a guarantee for maximum probability of a type I or a type II error is the target of the test, then, by definition, p-values must be computed, and their dependency on the stopping intention and testing specification must be considered~\cite{Kruschke.2018}. %

\textit{Assessment~(see~\tableref{tab:RABayesianStatistics}): }
\begin{table}[bpt]
    \centering
    \setlength\tabcolsep{4.5pt}
    \caption{Assessment of \RABayesianStatistics{}}
    \begin{tabularx}{\columnwidth}{|c|c|c|C|}
    \hline
    \CritQuantifiablityShort{} & \CritValidityShort{} & \CritMissingLinksShort{} & \CritBlackboxShort{}\\ \hline
    \Positive & \Negative{} ... \Positive & \Negative{} ... \Positive & \Positive
 \\ \hline
    \end{tabularx}
    \label{tab:RABayesianStatistics}
\end{table}
We conclude that the choice of Bayesian statistics alone does not justify a reduction of the real-world driving requirement. %
However, Bayesian statistics offers a more robust interpretation of results and~\textemdash{}~in theory~\textemdash{}~enables the integration of diverse information types, such as expert knowledge and pre-test data as stated in~\cite{Siu.1998}. %
It provides a mathematically \CritQuantifiablity[quantifiable]{} mechanism for the reduction of the real-world driving requirement using "knowledge from prior vehicle programmes"~\cite[52]{SOTIF2022} or "using expert knowledge with similar systems"~\cite[156]{SOTIF2022}. %
However, no standardized method currently exists for incorporating such input into the prior definition without introducing uncertainties that cannot be reliably measured or objectively quantified. %
This represents a \CritMissingLinks[missing link]{} and, if unaddressed, may pose significant \CritValidityshort{}. %
Consequently, the ratings for both categories vary~\textemdash{}~from negative to positive~\textemdash{}~depending on the rationale for the prior and the assumptions and methods used in its definition. %
Another \CritValidity[threat to validity]{} arises when incorporating knowledge from previous systems, since software is susceptible to changes~\cite{Madala.2022} and even small changes in system architecture or functionality can lead to significant shifts in behavior, making prior insights potentially unrepresentative. %

In general, the Bayesian approach is quantifiable and well-documented. %
However, until a standardized universally accepted method for the definition of a prior under incorporation of such additional information is available, we recommend adherence to the \textit{principle of indifference}, given the strong influence of the prior on the outcome. %
When following this principle, \textit{Bayesian statistics} is typically \CritBlackbox[black-box compatible]{}, relying solely on the statistical properties of the observed data and the defined assumptions. %

\RCSubSection{\RCUncertainty{}}\label{sec::RCUncertainty}%
The \glspl{RA} described in this section share the commonality that they achieve test reduction by enhancing the efficiency of the estimation process, thus allowing equivalent or more informed statements about the frequency of occurrence of \gls{HB} with a smaller database. 
\RPSubSubSection{\RPTestProfile{}}\label{sec::RPTestProfile}%
To obtain results that reflect operation of the \gls{SuT} within the target \gls{ODD}, \gls{FOT} must expose the \gls{SuT} to all relevant stimuli~\textemdash{}~such as scenarios and environmental conditions~\textemdash{}~at frequencies that match real-world deployment of the \gls{SuT}.
The rationale of this \gls{RP}, however, is to improve the efficiency of parameter estimation by deliberately refining the test profile such that variance in the data is reduced. %
Since this refinement deviates from the baseline distribution of stimuli, the resulting estimates may no longer be representative, and a bias correction must be applied to restore their validity. %
\RAParagraph{\RAHighYieldRouteProfiles{}}\label{sec::RAHighYieldRouteProfiles}
According to \citenumber{SOTIF2022}, such a refinement could be achieved by omitting testing conditions under which the \gls{SuT} is inactive or by selecting test routes for \gls{FOT} based on their criticality~\cite[52]{SOTIF2022}. %
\citeauthor{Glauner.2012}~\cite{Glauner.2012} propose a concept for the efficient \gls{FOT} of \gls{ADAS} systems that reflects this idea. %
They propose an iterative process in which data concerning the occurrence of events of interest is extracted from \gls{FOT} and assessed regarding its relevance to a given test target. %
This assessment is then linked to the specific location or circumstances of the event occurrence, resulting in different weights being assigned to specific road sections. %
Once specified, these weights enable the iterative planning of high yield route profiles. %
These ideas are further detailed and practically applied for the automated planning of a high yield route profile in~\cite{Glauner.2014}.  %
A key aspect highlighted in the outlooks of~\cite{Glauner.2012, Glauner.2014} is the need for a quantification of the effect, which is seen as a topic for future research. %

A potential solution to this shortcoming regarding the quantifiability of the effect is the mathematical principle of \gls{IS}, which is commonly used in studies investigating the \hyperref[sec::RASBT]{\gls{SBT}} of \gls{ADS}~\cite{Zhao.2017c, Gelder.2017, Zhao.2018, Jesenski.2020, Jesenski.2021}. %
\gls{IS} is a technique used to efficiently estimate statistical parameters, particularly when the event of interest is rare under the original distribution. %
Instead of sampling directly from the original distribution, samples are drawn  from another distribution that overrepresents the rare event, referred to as the \gls{IS} distribution. %
To correct for the bias introduced by this technique, each sample is weighted by the likelihood ratio describing the ratio of the original density to the density of the more efficient \gls{IS} distribution. %
The main challenge in the application of \gls{IS} is the selection of an appropriate \gls{IS} distribution, particularly when the parameter space is high-dimensional~\cite{Jesenski.2020}. %
This difficulty becomes pronounced in \gls{FOT}, where a route may be described by a multitude of parameters~\textemdash{}~ranging from geometric characteristics to environmental and traffic conditions~\textemdash{}~that can individually and collectively affect the behavior of complex \gls{ADAS} functions like \gls{AEB}. %

\figureref{fig::Extension_RAHighYieldRouteProfiles} illustrates the approach of \RAHighYieldRouteProfiles{} in combination with the mathematical foundation of \gls{IS}. %
Given a specified baseline distribution of parameters characterizing the test profile for \gls{FOT}, the distribution is skewed towards constellations that enhance the probability of occurrence of \gls{HB}, resulting in a high yield test profile. %
The bias of the test observation is then corrected according to the given baseline. %
This corrected observation is subsequently used for test evaluation. %

\textit{Assessment~(see~\tableref{tab:RAHighYieldRouteProfiles}): }
\begin{figure*}[!hbtp]
\tikzsetnextfilename{Fig_12_Reduction_Approach_High_Yield_Route_Profiles}
\centering
\includegraphics[]{pictures/Base_Model/RAHighYieldRouteProfiles.tikz}
\caption{\FigApproachPrefix{}\RAHighYieldRouteProfiles{}}
\label{fig::Extension_RAHighYieldRouteProfiles}
\end{figure*}
While the principle of challenging a \gls{SuT} by specification of challenging test profiles in \gls{FOT} is mentioned in the literature, we were not able to locate literature that attempts to reason a quantified reduction of the real-world driving requirement based on especially challenging test profiles. %
However, a comparison with literature on statistical inference from \gls{SBT} (see \hyperref[sec::RASBT]{\textit{RA: \RASBT{}}}) suggests that such an approach is \CritQuantifiablity[potentially quantifiable]{} by the mathematical principles of \gls{IS}. %
Since no approaches implementing \gls{FOT} in conjunction with \gls{IS} were located in the literature, this concept remains hypothetical and its practical applicability remains uncertain, which is reflected in the corresponding assessment. %
Accordingly, we see a major \CritMissingLinks[missing link]{} in the modelling of distributions describing specific \gls{FOT} profiles due to the huge potential parameter space regarding this use case. %
Furthermore, even if such distributions can be found for a given \gls{SuT}, they may not necessarily be transferable to other \gls{SuT} implementations and/ or applications. %
If a statistically significant derivation of these parameter distributions for each individual \gls{SuT} is possible however, this process constitutes an additional effort that relies on \gls{FOT} data, which counteracts the potential test effort reduction. %
Moreover, the \CritValidity[validity]{} of this approach is threatened by the challenge of distinguishing causality from correlation in the search for efficient \gls{FOT} test profiles: %
If the real-world driving requirement is reduced based on mere correlations between system failures and the chosen route profile without the establishment of a causal link, the validity of the results becomes questionable. %
A notable exception to these threats to validity is the omission of test conditions under which the system is guaranteed to be inactive. %
Also, the approach is \CritBlackbox[black-box compatible]{}, even though a grey-box or white-box model of the \gls{SuT} would certainly be useful for the identification of suitable route or test profiles. %
\begin{table}[tbp]
    \centering
    \setlength\tabcolsep{4.5pt}
    \caption{Assessment of \RAHighYieldRouteProfiles{}}
    \begin{tabularx}{\columnwidth}{|c|c|c|C|}
    \hline
    \CritQuantifiablityShort{} & \CritValidityShort{} & \CritMissingLinksShort{} & \CritBlackboxShort{}\\ \hline
    \Negative{}~... \Positive & \Negative{} & \Negative{} & \Neutral{}~... \Positive
 \\ \hline
    \end{tabularx}
    \label{tab:RAHighYieldRouteProfiles}
\end{table}

\RPSubSubSection{\RPValueEstimation{}}\label{sec::RPValueEstimation}%
This \gls{RP} focuses on enhancing the estimation quality of the frequency of the \gls{HB} without alteration of the test profile, leveraging alternative methods.%
\RAParagraph{\RAEVA{}}\label{sec::RAEVA}
\gls{EVT} is a statistical field that models the probability of occurrence of \textit{extreme} events by specialized probability distributions. %
In this context, \textit{extreme} refers to values that are especially high or low in magnitude in relation to the values frequently observed for this statistical variable. %

\AA{}sljung et al. applied \gls{EVA} to the results of longitudinal criticality metrics in naturalistic driving data to predict the probability of collisions in the field~\cite{Asljung.2016, Asljung.2017, Asljung.2018}. %
They achieved results consistent with applicable traffic statistics data and significantly outperformed the \gls{NHST} with respect to the driving distance required. %
However, the results also indicate that the accuracy of the collision probability estimates highly depends on the choice of criticality metric and is sensitive to variations in degrees of freedom inherent to the \gls{EVA}. %
In~\cite{Asljung.2022} and~\cite{Gyllenhammar.2023}, similar approaches are proposed for the validation of the collision frequency of \gls{ADS} and extended further towards predictive monitoring of \gls{ADS}.
Approaches that extend this idea to the application to the validation of \gls{ADAS} are also present in the literature. %
\citeauthor{Stellet.2024}~\cite{Stellet.2024} describe a generic approach for safeguarding a system against \gls{FP}, which includes the proposal to leverage \gls{EVA} in the process. %
The approach is based on mapping a criticality assessment of a system and a given reference criticality to a generic error measure referred to as \gls{TPI}. %
They propose to employ \gls{EVA} of the \gls{TPI} values to calculate the \gls{FP} rate of a given \gls{SuT}, suggesting an application of the approach to \gls{AEB} systems. %
\citeauthor{Schrimpf.2024}~\cite{Schrimpf.2024} propose a similar approach for the application to \gls{ADAS}, more specifically emergency intervention systems using a different metric for the \gls{EVA}, which is referred to as the \gls{IPM}. %
Based on an assessment of the correctness of the system behavior, they propose to use the distribution of \gls{IPM} values associated with incorrect behavior of the \gls{SuT} to estimate the \gls{FP} rate of the \gls{SuT} using \gls{EVA}. %
\figureref{fig::RAEVA} illustrates the principle of application of the \gls{EVT} according to the approaches described in this section. %
A \gls{PM} to the \gls{HB} is defined and applied to the data. %
Subsequently, an \gls{EVT} distribution model is fitted to the \gls{PM} values, resulting in a probability density function that can be used to derive an estimate of the probability of occurrence for extreme values of this \gls{PM}. %
In addition to laying the groundwork for an improved estimation, the application of a \gls{PM} to \gls{FOT} data also provides the benefit of providing additional insights into the system performance~\cite{Stellet.2024,Schrimpf.2024}.%
\begin{figure*}[btp]
\tikzsetnextfilename{Fig_14_Reduction_Approach_Substitution_of_Hypothesis_Testing_by_Extreme_Value_Analysis}
\centering
\includegraphics{pictures/Base_Model/RAEVA.tikz}
\caption{\FigApproachPrefix{}Substitution of Hypothesis Testing by \RAEVA{}}
\label{fig::RAEVA}
\end{figure*}

\textit{Assessment~(see~\tableref{tab:RAEVA}): }%
In summary, the principle of using \gls{EVA} to reduce the test effort for \gls{ADAS} seems very promising: The results of \gls{EVA} can be \CritQuantifiablity[quantified adequately]{} and outperform \glspl{NHST} in similar use cases in the literature~\cite{Asljung.2016, Asljung.2017, Asljung.2018}. %
Approaches that outline the application of \gls{EVA} to the given problem exist in the literature and have no major \CritMissingLinksshort{}. %
However, criteria for the specific design and implementation of some steps, such as selecting the most appropriate \gls{PM}, are not present in the literature. %
The only significant \CritValidity[threat to the validity]{} we identified is the validity of the underlying assumption that the mathematical prerequisites for the application of the \gls{EVT} are met in a given use case, which also depends on the choice of \gls{PM}. %
Due to the potential specificities of the implementation of the method for a given \gls{SuT} and the associated data, we were unable to assess this aspect generically. %
The existing approaches described in~\cite{Stellet.2024, Schrimpf.2024} are \CritBlackbox[not black-box compatible]{} because their application is based on the calculation of a \gls{PM} from internal system information. %
While a complete white-box model of the \gls{SuT} may not always be necessary, at least a gray-box model of the \gls{SuT} is inherently required. %
However, if the \gls{PM} is defined in terms of observable behavior, such as minimum distances to \gls{VRU}, for example, even a black-box model of the system can be sufficient. %
\begin{table}[tbp]
    \centering
    \setlength\tabcolsep{4.5pt}
    \caption{Assessment of \RAEVA{}}	
    \begin{tabularx}{\columnwidth}{|c|c|c|C|}
    \hline
    \CritQuantifiablityShort{} & \CritValidityShort{} & \CritMissingLinksShort{} & \CritBlackboxShort{}\\ \hline
    \Positive & \Negative{}~... \Positive & \Neutral & \Negative{} ... \Positive
 \\ \hline
    \end{tabularx}
    \label{tab:RAEVA}
\end{table}

\RCSubSection{\RCSupplement{}}\label{sec::RCSupplement}%
In contrast to the aforementioned approaches, the third cluster identified is based on the supplementation or the substitution of \gls{FOT} with alternative testing methods that inherently require less testing effort. %
\RPSubSubSection{\RPSimBT{}}\label{sec::RPSimBT}%
Simulative testing encompasses various levels of virtualization ranging from different forms of \gls{XiL} techniques~\cite{Riedmaier.2018} to complete virtualization. %
In comparison to \gls{FOT}, simulation offers significant benefits regarding time and cost efficiency as well as safety. %
Consequently, the complete or partial replacement of \gls{FOT} by simulative testing methods appears as an attractive option for test effort reduction. %
However, there is a central inherent disadvantage to this \gls{RP}: Virtualization potentially threatens the validity of assessment results. %
Therefore, the validity and completeness of simulation models are essential for the credibility of simulation results. %
Accordingly, complete virtualization is currently limited by the insufficient fidelity of sensor, environment, and vehicle models, which restricts the ability to replicate real-world complexity~\cite{Beringhoff.2022, Wachenfeld.2016, ISO.5083, Drechsler.2022}. %
Methods that supplement real-world data with virtual inputs~\textemdash{}~such as the mileage multiplier approach proposed in~\cite{Augustsson.2010}~\textemdash{}~inherently face the same validation issues as the simulation of the entire vehicle and its environment, since even minor discrepancies in model accuracy can propagate and affect simulation outcomes significantly~\cite{Drechsler.2022,Dona.2022}. %

\RAParagraph{\RAVirtualFOT{}}\label{sec::RAVirtualFOT}%
The Reuse of \gls{FOT} data for resimulation of software is a common industry practice in the validation and development of \gls{ADAS}~\cite{Krishnan.2022} and \gls{ADS}~\cite{appliedintuition.2021}. %
In this process, software components are tested based on a replay of real-world perception data as input. %
If the resimulation is validated and changes are limited to software, this technique eliminates the need for a separate \gls{FOT} for each software change during development. %
However, if hardware components~\textemdash{}~such as the sensor setup~\textemdash{}~are modified, the requirement for full-scale \gls{FOT} of the complete \gls{SuT} remains. %
This is because variations in sensor properties~\textemdash{}~such as resolution~\cite{Scheiner.2020} or mounting position~\cite{Elster.2023}~\textemdash{}~potentially impact detection performance metrics and the effectiveness of downstream functions~\cite{Lee.2021}. %
Furthermore, the dynamics and vibration behavior of the test vehicle may influence perception sensor measurements~\cite{Hau.2020}. %
Nevertheless, a reduction of the real-world driving requirement by reusing \gls{FOT} data is only possible with data that does not need to be recorded on the final hardware setup of the \gls{SuT}. %

While we were unable to identify a suitable approach to overcome the mentioned challenges in the scientific literature, an industry press release from 2021 claims the development of a system capable of transforming an existing \gls{FOT} dataset into sensor inputs "as seen" by new sensors in new vehicle applications~\cite{Sapienza.2021}, and advertises this concept as a substitute for \gls{FOT} of new systems. %
However, no detailed information on the implementation is publicly available. %
At the time of writing, the aforementioned press release is the only source of information on the concept that we were able to locate. %

\figureref{fig::RAVirtualFOT} illustrates the concept of reusing existing \gls{FOT} data for the validation of a new \gls{SuT} as postulated by~\cite{Sapienza.2021} as a substitute to a real \gls{FOT}. %
\begin{figure*}[!hbtp]
\tikzsetnextfilename{Fig_14_Reduction_Approach_Reuse_of_existing_FOT_Data_in_a_Virtual_FOT}
\centering
\includegraphics{pictures/Base_Model/RAVirtualFOT.tikz}
\caption{\FigApproachPrefix{}Reuse of existing \gls{FOT} Data in a \RAVirtualFOT{}}
\label{fig::RAVirtualFOT}
\end{figure*}
Existing data is modified to imitate data recordings produced by the \gls{SuT}. %
The software component of the \gls{SuT} is stimulated with a replay of the modified data, which creates an observation used to evaluate the achievement of the \validationtarget{}. %

\textit{Assessment~(see~\tableref{tab:RAVirtualFOT}): }
Reusing existing \gls{FOT} data as a supplement or replacement for \gls{FOT} in the validation process despite hardware changes is an appealing idea, as the reduction of the real-world driving requirement is \CritValidity[easily quantifiable in theory]{}. %
However, a method to assess the transferability of existing or synthetic data to new hardware including a corresponding validation is not reflected in the scientific literature even though at least one method has reportedly been developed in the industry~\cite{Sapienza.2021}. %
This constitutes a major \CritMissingLinks[missing link]{} for an application of the approach. %
An independent \CritValidity[threat to validity]{} exists in the fact that the traffic environment is not necessarily constant over time; so older data may be less relevant for the safety validation of a new system prior to its market introduction~\cite{Junietz.2018}. %
Because this approach incorporates a complete simulation of the \gls{SuT}, it is inherently \CritBlackbox[not black-box compatible]{}. %
\begin{table}[bp]
    \centering
    \setlength\tabcolsep{4.5pt}
    \caption{Assessment of \RAVirtualFOT{}}
    \begin{tabularx}{\columnwidth}{|c|c|c|C|}
    \hline
    \CritQuantifiablityShort{} & \CritValidityShort{} & \CritMissingLinksShort{} & \CritBlackboxShort{}\\ \hline
    \Positive & \Neutral & \Negative & \Negative
 \\ \hline
    \end{tabularx}
    \label{tab:RAVirtualFOT}
\end{table}

\RPSubSubSection{\RPPGT{}} \label{sec::RPPGT}
\gls{PGT} delivers high-fidelity, real-world data in a controlled setting, yielding realistic data for assessing system behavior and making it an attractive supplement to \gls{FOT}.
The main benefit is that \gls{PGT} allows for reproducing rare traffic events, allowing for a more rapid assessment of the use-case performance of, e.g., emergency intervention systems~\cite{ISO.5083,SaFAD.2019}. %
This controlled approach enhances safety and reproducibility compared to \gls{FOT}, but incurs significant costs and operational demands, such as additional setup time and financial resources~\cite{Beglerovic.2017}. %
The main drawback of \gls{PGT} is the strongly simplified environment compared to \gls{FOT}~\cite{Wachenfeld.2016,King.2020}. %
Another major challenge is the orchestration of complex scenarios since precise control of all actors is essential to achieve repeatability~\cite{Beringhoff.2022}. %
In addition, the inherent non-replicability of environmental conditions poses a challenge for \gls{SBT} on proving grounds. %

\RAParagraph{\RASBT{}}\label{sec::RASBT}%
\gls{SBT} is widely accepted as a more effective and efficient solution than \gls{FOT} to the validation challenge of \gls{ADS}~\cite{Khastgir.2021, Amersbach.2019, Weissensteiner.2023, Neurohr.2020, Gelder.2021} and \gls{ADAS}~\cite{Zhang.2023} in the scientific community and is an established concept in normative~\cite{SOTIF2022} and regulatory~\cite{UNECE.06.04.2023,UNECE.04.03.2021} contexts. %
\citeauthor{Menzel.2018}~\cite{Menzel.2018} differentiate between three levels of abstraction for scenario description: \textit{Functional scenarios} are descriptions in natural language, \textit{logical scenarios} are parametrized scenario representations and \textit{concrete scenarios} are specific instances of logical scenarios with specific values assigned to all parameters. %
Potential testing platforms for the execution of concrete scenarios range from proving ground testing and \gls{XiL}-approaches~\cite{Riedmaier.2018} to complete virtualization of the \gls{SuT} and the environment via simulation tools. %
In contrast to \gls{FOT}, there is no direct quantitative link of \gls{SBT} to an estimated mileage~\cite{Junietz.2018} and thus to the real-world driving requirement, which creates a challenge in \gls{QSVRR} when \gls{SBT} is used as a supplement to or substitute for \gls{FOT}. %
A quantifiable supplementation or substitution of \gls{FOT} by \gls{SBT} requires that the scenarios chosen achieve at least the same \textit{coverage} as a \gls{FOT}. %
For this purpose, suitable coverage metrics are necessary. %
An existing approach~\cite{Amersbach.2019} estimates the equivalent mileage for a given number of scenarios based on estimating the distance that is driven in an average scenario and conservative assumptions about scenario overlap and the uniqueness of driving conditions~\cite{Langner.2018} encountered in naturalistic driving. %
While this approach can serve as a starting point, it only derives a number of scenarios to be tested while not specifying  which scenarios should be tested. %
\citeauthor{FlorianHauer.2019} present a method for data-driven scenario elicitation that addresses this aspect. The method entails the quantification of the probability that a new scenario type with a given probability of occurrence has not yet been discovered in a set of driving data~\cite{FlorianHauer.2019}. %
De \citeauthor{Gelder.2024} present a metric for the quantification of the completeness of \textit{activities} in a given data set which constitute the building blocks for scenarios~\cite{Gelder.2019}. %
In a different publication, different metrics to quantify the extent to which a given collection of scenarios covers a given \gls{ODD} and a given data set are presented~\cite{Gelder.2024}. %
A remaining challenge that coverage metrics alone cannot solve is the fact that achieving full coverage might not always be guaranteed upon collecting infinitely more data due to inadequate data acquisition strategies~\cite{FlorianHauer.2019} or even possible at all~\cite{Gelder.2024, Zhang.2023}. %
Another challenge in the application of \gls{SBT} is scenario parametrization. %
The process of parametrization can be decomposed into two separate steps: parameter space discretization and parameter sampling. %
The choice of discretization granularity constitutes a tradeoff: A finer discretization results in more test cases and thus more testing effort while a coarse discretization leads to less parameter space coverage~\cite{Amersbach.2019}. %
An important assumption for parameter space discretization is that properties of interest, such as trajectories, are smooth between discretization steps. %
If proven valid, this assumption increases the credibility of the transferability of a finite set of concrete scenario simulations to the continuous parameter space~\cite{Neurohr.2020}. %

De \citeauthor{Gelder.2021} present a concept for the quantitative risk assessment of an \gls{ADS} by simulative \gls{SBT}~\cite{Gelder.2021} that is also applicable to \gls{ADAS}. %
In this approach, the scenario parametrization is based on a data-driven estimation of the exposure of the \gls{SuT} to specific scenario parameters. %
In condensed and generalized form, the approach encompasses the following steps: %
\begin{enumerate}
    \item Identify the scenarios that the \gls{ODD} of the \gls{SuT} contains.
    \item Determine the exposure of the \gls{SuT} to each scenario in terms of the expected number of occurrences per hour of driving.
    \item Simulate the response of the \gls{SuT} in these scenarios.
    \item Calculate the expected number of undesired events.
\end{enumerate}
The approach is illustrated with a case study using an example system and corresponding exemplary scenario categories. %
One of the key conclusions is that more research is required to identify the relevant scenario categories for a complete risk assessment of a given \gls{SuT}~\cite{Gelder.2021}. %

\figureref{fig::RASBT} summarizes the individual steps associated with \gls{SBT} and a substitution of \gls{FOT} by \gls{SBT} as discussed in the previous section, centred around the approach of de~\citeauthor{Gelder.2021}.~\cite{Gelder.2021} mentioned above. %
\begin{figure*}[btp]
\tikzsetnextfilename{Fig_15_Reduction_Approach_Substitution_of_FOT_by_SBT}
\centering
\includegraphics{pictures/Base_Model/RASBT.tikz}
\caption{\FigApproachPrefix{}Substitution of \gls{FOT} by \RASBT{}}
\label{fig::RASBT}
\end{figure*}
\textit{Logical scenarios} that require testing are elicited and \textit{concrete scenarios} are derived using suitable parameter discretization and sampling methods. %
Subsequently, an observation is generated by executing \textit{concrete scenarios} on a chosen test platform. %
The observed number of \gls{HB} is translated into the equivalent expected value of \gls{HB} in natural traffic based on the exposure to the tested scenarios. %
The result of this translation serves as the comparative value to the \validationtarget{}, which is derived on the premise of \gls{FOT} as a testing method. %

\textit{Assessment~(see~\tableref{tab:RASBT}): }
\begin{table}[btp]
    \centering
    \setlength\tabcolsep{4.5pt}
    \caption{Assessment of \RASBT{}}
    \begin{tabularx}{\columnwidth}{|c|c|c|C|}
    \hline
    \CritQuantifiablityShort{} & \CritValidityShort{} & \CritMissingLinksShort{} & \CritBlackboxShort{}\\ \hline
    \Neutral~... \Positive & \Neutral & \Neutral & \Negative~... \Positive
 \\ \hline
    \end{tabularx}
    \label{tab:RASBT}
\end{table}
In summary, conceptual approaches to replace distance-based testing by \gls{SBT} are present in the literature. %
However, it appears that the transition to \textit{macroscopic} safety assessments seems to be largely overlooked in the otherwise rich literature on \gls{SBT}~\cite{Riedmaier.2020}. %
This deficiency becomes especially apparent concerning the direct comparability of \gls{SBT} to \gls{FOT} mileages~\cite{Beringhoff.2022} and thus the \CritQuantifiablity[quantifiability]{} of \gls{SBT}. %
In general, further research is required for the practical application of published approaches~\cite{Riedmaier.2020}. %
For example, while the coverage metrics mentioned above already provide mathematical tools for the quantification of the completeness of scenarios or activities with respect to different references, the definition of completeness requirements~\textemdash{}~i.e., coverage thresholds~\textemdash{}~is required for practical application. %
However, we were unable to locate specific coverage threshold values in the literature. %
While there is not a specific, singular \CritMissingLinks[missing link]{} preventing the application of \gls{SBT} for a reduction of the real-world driving requirement, a reduction of the overall test effort by \gls{SBT} in practical applications is effectively uncertain~\cite{Neurohr.2020} and yet to be proven. %
This lack of a "proof of industrialization" of \gls{SBT} casts doubt on the viability of \gls{SBT} in this context since the complexity and scope of application examples in the literature are typically limited~\cite{Riedmaier.2020}. %
This gap is reflected in the uncertainty about the practical application of simulative \gls{SBT} in the industry~\cite{Beringhoff.2022}. %
Furthermore, \gls{SBT} faces inherent and unavoidable \CritValidityshort{}: It is probably impossible to formulate a complete scenario description that would be able to cover every effect or influence that may be present in the real world~\cite{Junietz.2018}. %
Additionally, it has been shown that the requirement for the properties of interest to be smooth between discretization steps does not generally hold~\cite{Mori.2023}. %
In a related survey, no approach for the assurance of this property could be located in the literature~\cite{Neurohr.2020}. %
Finally, \gls{SBT} inherits threats to validity related to the testing platform chosen as described in sections \textit{Simulation-based Testing} and \textit{Proving Ground Testing}. %
The \CritBlackbox{} of \gls{SBT} is determined by the choice of testing platform: Proving ground tests with real vehicles are black-box compatible without limitations, while different levels of virtualization require different levels of insight into the \gls{SuT} up to a white-box model for complete virtualization. %

\section{Summary}\label{sec::Summary}%
This publication systematically addresses approaches regarding the reduction of the real-world driving requirement of \mbox{\gls{FOT}-based} \gls{QSVRR} for \gls{ADAS} and \gls{ADS}. %
Based on an analysis of \citenumber{SOTIF2022}, the \genericmodel{}{}, which reflects the key input parameters and intermediate stages for the derivation of a generic \validationtarget{}, is introduced. %
Using an \gls{AEB} functionality as an example, the \basemodel{} is derived, which establishes a specific baseline for the derivation of a real-world driving requirement based on the \genericmodel{}. %
The scientific literature is reviewed concerning potential \glspl{RA} for the established real-world driving requirement that are compatible with the \genericmodel{} and the \basemodel{}. The findings of the literature review are categorized into a hierarchy of distinct \glspl{RC}, \glspl{RP} and \glspl{RA} based on the underlying mechanisms. %
Every \gls{RA} identified from the literature review is presented in detail, culminating in assessments of each approach regarding the criteria \CritQuantifiablity{}, \CritValidity{}, \CritMissingLinks{}, and \CritBlackbox{}. %
Finally, the implications of the findings for the future of the safety validation of \gls{ADAS} and \gls{ADS} are discussed. %

\section{Conclusions}\label{sec::Conclusion}%
Numerous \glspl{RA} were identified and systematized as a result of the literature review. %
As the described categorization of \glspl{RA} into different \glspl{RC} and \glspl{RP} already suggests, the presented approaches are not exhaustive. Rather, additional \glspl{RA} may be derived from the existing structure~\textemdash~such as \hyperref[sec::RPValueEstimation]{\textit{\RPValueEstimation{}}}~\textemdash~which could represent an alternative estimation method to \hyperref[sec::RAEVA]{\textit{\RAEVA{}}}. %
It is also conceivable to combine multiple \glspl{RA} within a single \gls{QSVRR} implementation. %
For example, a \hyperref[sec::RAVirtualFOT]{\textit{\RAVirtualFOT{}}} conducted using \hyperref[sec::RAHighYieldRouteProfiles]{\textit{\RAHighYieldRouteProfiles{}}} in conjunction with \hyperref[sec::RADecompositionHB]{\textit{\RADecompositionHB{}}} would be a plausible combination. %

However, what unites all of the discussed \glspl{RA} is the fact that none of the approaches is broadly accepted for \gls{QSVRR} and fully formalized, e.g., in terms of requirements to uphold validity. %
Some \glspl{RA} are free of major missing links, like \hyperref[sec::RADecompositionAH]{\textit{\RADecompositionAH{}}}, \hyperref[sec::RASequentialTesting]{\textit{\RASequentialTesting{}}}, and \hyperref[sec::RASequentialTesting]{\textit{\RARedundancy{}}} but still lack guidance and examples for practical implementation. %
For other \glspl{RA} that were identified, further research is required to realize their full potential and enable practical applications: %
\begin{itemize}
\item The \hyperref[sec::RADecompositionHB]{\textit{\RADecompositionHB{}}} requires methods to ensure validity for the estimation of the conditional probabilities when system dependence cannot be excluded.
\item The potential of \hyperref[sec::RABayesianStatistics]{\textit{\RABayesianStatistics{}}} is currently limited by the absence of robust mechanisms for integrating expert knowledge or historical system data into the prior definition.
\item To justify test reductions based-on \hyperref[sec::RAHighYieldRouteProfiles]{\textit{\RAHighYieldRouteProfiles{}}}, methods for translating biased results into unbiased equivalents need to be developed, potentially employing concepts like \gls{IS}, and for preventing purely correlation-based reductions.
\item For \hyperref[sec::RAEVA]{\textit{\RAEVA{}}}, further research must establish clear requirements and guidelines for the definition of suitable \glspl{PM} and the application of \gls{EVA}.
\item \hyperref[sec::RAVirtualFOT]{\textit{\RAVirtualFOT{}}} requires validated methodologies to assess the transferability of existing or synthetic datasets to new hardware setups to justify substitution or supplementation.
\item \hyperref[sec::RASBT]{\textit{\RASBT{}}} requires broadly accepted coverage metrics and thresholds that quantify sufficient coverage. Additionally, its suitability as a direct substitute remains uncertain due to intrinsic limitations in scenario description and associated testing platforms like \hyperref[sec::RPPGT]{\textit{\RPPGT{}}} and \hyperref[sec::RPSimBT]{\textit{\RPSimBT{}}}.
\end{itemize}
Among these \glspl{RA}, \we{} consider research on those as particularly important that analyze the precursors of the \gls{HB} under assessment, like \hyperref[sec::RADecompositionHB]{\textit{\RADecompositionHB{}}} and \hyperref[sec::RAEVA]{\textit{\RAEVA{}}}. %
Throughout these analyses, these approaches generate additional knowledge about internal system behavior, making them assets for safety \gls{V&V} beyond the scope of purely \gls{QSVRR}-oriented approaches. %

Furthermore, we want to highlight that the scientific community seems to widely agree that \gls{SBT} will significantly shape the future of \gls{V&V} for \gls{ADS} and \gls{ADAS}. %
However, even if the unresolved challenges and inherent threats to validity of this approach are overcome in the future, a residual dependence on \gls{FOT} is likely to remain, due to the data-driven nature of coverage verification~\textemdash{}~and potentially the scenario elicitation process~\textemdash{}~in the implementations documented in the literature. %

Regardless of how the safety validation of \gls{ADAS} and \gls{ADS} will be implemented in the future, the literature review reveals a significant gap: aside from a few specific distance calculations in the context of distance-based testing, specific and detailed \gls{QSVRR} implementations are largely absent from current literature. %
Most literature addresses the relevant concepts and considerations only at an abstract level, which, from our perspective, is an insufficient state of the art to support the development of specific \gls{QSVRR} implementations. %
Addressing this gap requires significant future research and standardization~\textemdash{}~a conclusion reinforced by the analyses conducted in this paper. %

\printbibliography
\begin{IEEEbiography}[{\includegraphics[height=1.25in,width=1in,clip,keepaspectratio]{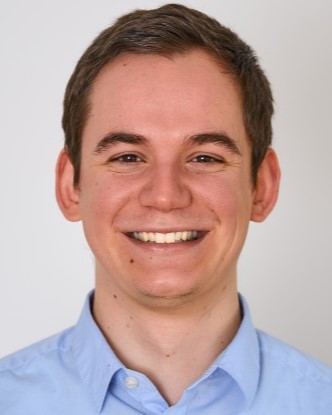}}]{Daniel Betschinske} was born in 1995. He received the B.\,Sc. degrees in mechanical and process engineering (2017) and in computational engineering (2021), and the M.\,Sc. degree in mechanical and process engineering (2022), all from \gls{TUDa}, Germany. Since February 2022, he has been a Research Associate at \gls{FZD} at \gls{TUDa}, where he devises methods utilizing prediction-centered retrospective analysis for efficient and standard-compliant \gls{ADAS} validation.\end{IEEEbiography}
\begin{IEEEbiography}[{\includegraphics[width=1in,height=1.25in,clip,keepaspectratio]{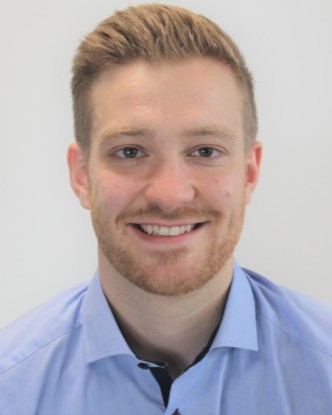}}]{Malte Schrimpf} was born in 1996. He received the B.\,Sc. and M.\,Sc. degrees in mechatronics from the \acrfull{TUDa} in 2018 and 2021 respectively. Since February 2022, he has been working as a a research associate at the \acrfull{FZD} at \gls{TUDa}, where he researches methods to increase the efficiency and effectiveness of \gls{ADAS} validation, focusing on \gls{FOT}-based methods in conjunction with extreme value analysis.\end{IEEEbiography}
\begin{IEEEbiography}[{\includegraphics[width=1in,height=1.25in,clip,keepaspectratio]{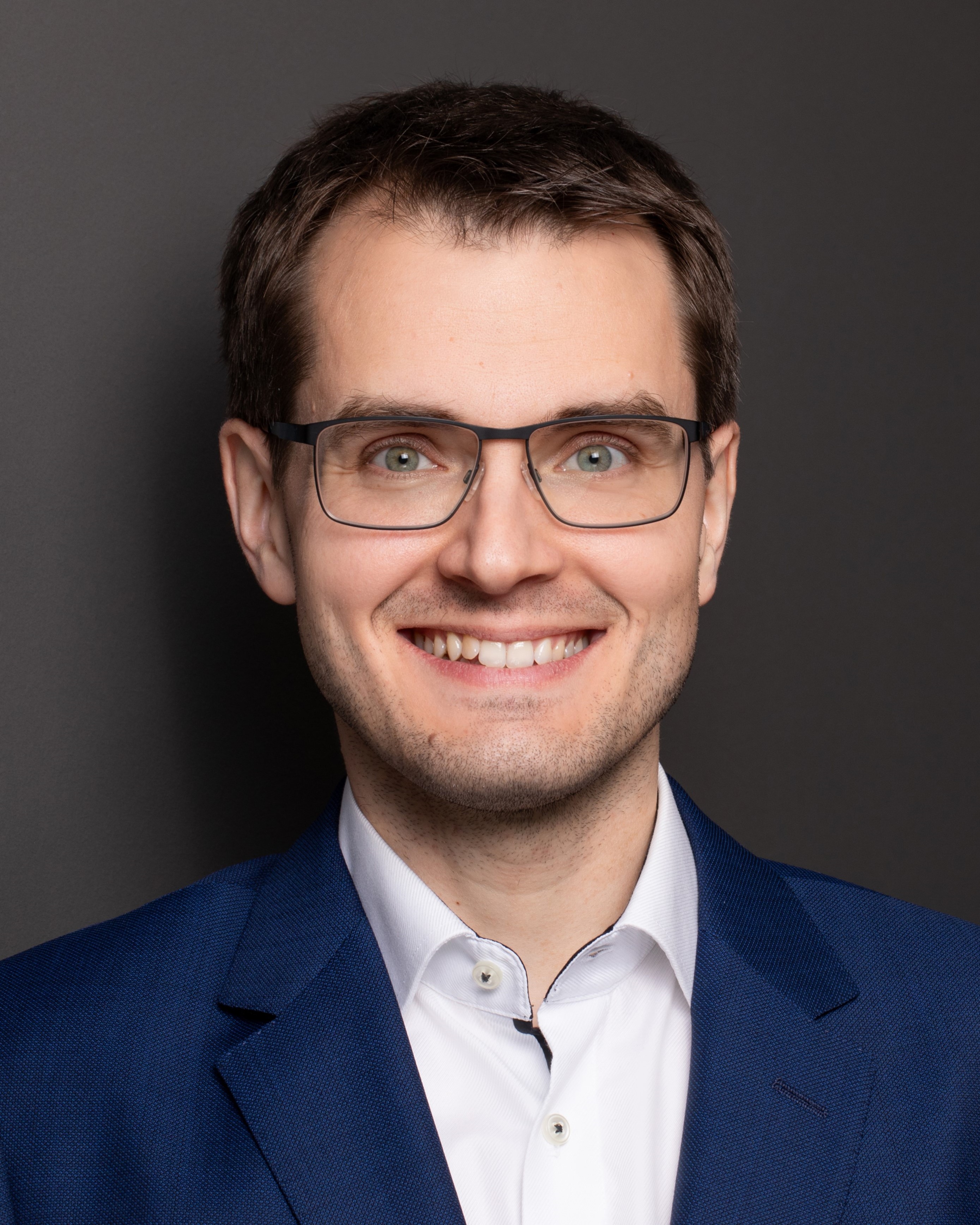}}]{Steven Peters} was born in 1987. He received a Dr.-Ing. degree from Karlsruhe Institute of Technology, Karlsruhe, Germany, in 2013. From 2016 to 2022, he was Manager of Artificial Intelligence Research with Mercedes-Benz AG, Germany. He has been a Full Professor with the \acrfull{TUDa} and heads the \acrfull{FZD}, Department of Mechanical Engineering, since 2022.\end{IEEEbiography}
\begin{IEEEbiography}[{\includegraphics[width=1in,height=1.25in,clip,keepaspectratio]{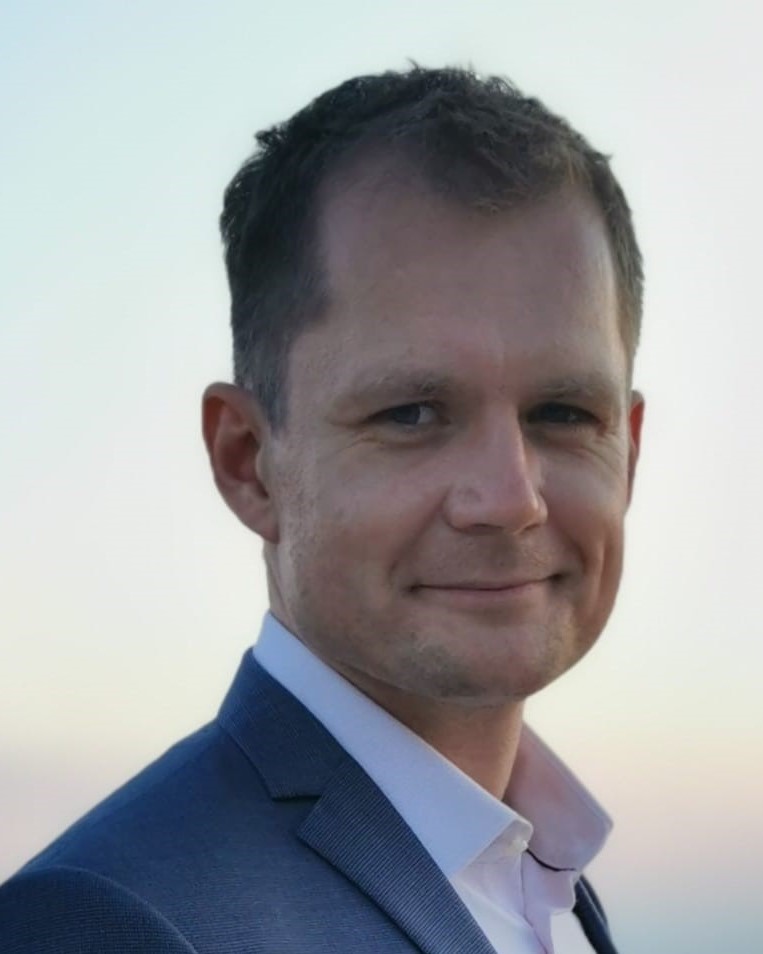}}]{Kamil Klonecki} was born in 1987. He received the B.\,Sc., M.\,Sc., and Dr.-Ing. degrees in mechanical and process engineering from the \acrfull{TUDa}, in 2010, 2013, and 2017, respectively.
He joined Continental AG, Frankfurt, Germany, in 2016. He is currently the Head of Data-Driven Development in the Business Area Autonomous Mobility at Continental AG.\end{IEEEbiography}
\begin{IEEEbiography}[{\includegraphics[width=1in,height=1.25in,clip,keepaspectratio]{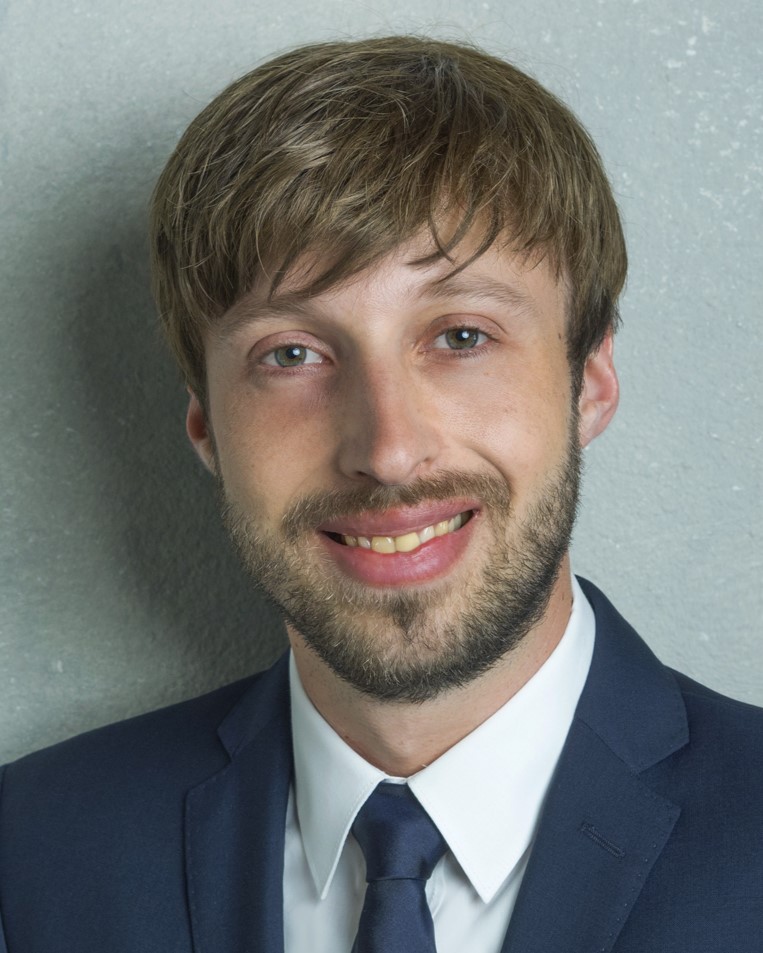}}]{Jan Peter Karch} completed his physics studies in 2011 and earned his Dr.\,rer.\,nat. degree in 2017, all at Johannes Gutenberg University, Germany.
Since 2018, he has been working at Continental Autonomous Germany GmbH in the Test \& Validation department for ADAS components.
As an Expert for Test Data Analytics, he is responsible for the test strategy for system performance and the conception of field operational tests.
\end{IEEEbiography}
\begin{IEEEbiography}[{\includegraphics[width=1in,height=1.25in,clip,keepaspectratio]{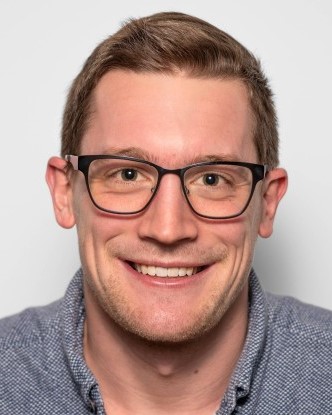}}]{Moritz Lippert} was born in 1993. He received the B.\,Sc. and M.\,Sc. degrees in mechanical and process engineering from \acrfull{TUDa} and the Dr.-Ing. degree from the \acrfull{FZD} at \gls{TUDa} in 2023. In his main research topic, the safety of automated driving, he investigated the derivation of scenery-based requirements for automated driving tasks and capability-based route planning.\end{IEEEbiography}
\end{document}